\documentclass[review]{elsarticle}
\makeatletter
\def\ps@pprintTitle{%
 \let\@oddhead\@empty
 \let\@evenhead\@empty
 \def\@oddfoot{\centerline{\thepage}}%
 \let\@evenfoot\@oddfoot}
\makeatother
\usepackage[top=1in, bottom=1.2in, left=1in, right=1in]{geometry}

\usepackage{hyperref}

\usepackage[ruled,linesnumbered,vlined]{algorithm2e}
\SetKwProg{Init}{initialization }{}{}
\usepackage{graphics,graphicx}
\usepackage{multicol} 
\usepackage{parskip}
\usepackage{amsmath}
\usepackage{multirow}
\usepackage[english]{babel}
\usepackage[utf8]{inputenc}
\usepackage{fancyhdr}
\usepackage{lipsum}
\usepackage[title]{appendix}
\usepackage{wasysym}
\usepackage{url}
\usepackage{adjustbox}
\usepackage{float}
\usepackage{caption}
\usepackage{subcaption}
\usepackage{pifont}%
\newcommand{\cmark}{\ding{51}}%
\newcommand{\xmark}{\ding{55}}%
\usepackage{cleveref}
\usepackage{tikz}
\usepackage{pgfplots}
\pgfplotsset{compat=newest}
\usepackage{amssymb, amsmath}
\usepackage{tikz}
\usepackage{ulem}
\usepackage{xcolor}
\usetikzlibrary{shapes, arrows, shapes.arrows,arrows.meta}
\usepackage[font=footnotesize,labelfont=small]{caption}
\usepackage[font=footnotesize,labelfont=small]{caption}

\begin{document}

\begin{frontmatter}

\title{Predicting Survival Outcomes in the Presence of Unlabeled Data}

\author[add1,add2]{Fateme Nateghi Haredasht\corref{c1}}
\ead{fateme.nateghi@kuleuven.be}
\author[add1,add2]{Celine Vens}
\address[add1]{KU Leuven, Campus KULAK - Department of Public Health and Primary Care, Etienne Sabbelaan 53, 8500 Kortrijk, Belgium}
\address[add2]{ITEC - imec and KU Leuven, Etienne Sabbelaan 51, 8500 Kortrijk, Belgium}
\cortext[c1]{Corresponding author}
\ead{celine.vens@kuleuven.be}

\begin{abstract}
Many clinical studies require the follow-up of patients over time. This is challenging: apart from frequently observed drop-out, there are often also organizational and financial challenges, which can lead to reduced data collection and, in turn, can complicate subsequent analyses. In contrast, there is often plenty of baseline data available of patients with similar characteristics and background information, e.g., from patients that fall outside the study time window. In this article, we investigate whether we can benefit from the inclusion of such unlabeled data instances to predict accurate survival times. In other words, we introduce a third level of supervision in the context of survival analysis, apart from fully observed and censored instances, we also include unlabeled instances. We propose three approaches to deal with this novel setting and provide an empirical comparison over fifteen real-life clinical and gene expression survival datasets. Our results demonstrate that all approaches are able to increase the predictive performance over independent test data. We also show that integrating the partial supervision provided by censored data in a semi-supervised wrapper approach generally provides the best results, often achieving high improvements, compared to not using unlabeled data.

\end{abstract}

\begin{keyword}
\text{Survival analysis}\sep Semi-supervised learning\sep Random survival forest \sep Self-training
\end{keyword}

\end{frontmatter}

\section{Introduction}\label{sec:intro}
Many clinical studies require following subjects over time and measuring the time until a certain event is experienced (e.g., death, progression, hospital discharge, etc). The resulting collected datasets are typically analyzed with survival analysis techniques.
Survival analysis is a branch of statistics that analyzes the expected duration until an event of interest occurs \cite{kleinbaum2010survival}. Censoring is an essential concept in survival analysis which makes it challenging compared to other analytical methods. Censoring can occur due to various reasons, such as drop-out, and means that the observed time is different from the actual event time. In the case of right censoring, for instance, we know that the actual event time is greater than the observed time \cite{hosmer2011applied}.     

Traditional survival analysis methods include the Cox Proportional Hazards model (CPH) \cite{cox1992regression}. CPH is basically a linear regression model that predicts simultaneously the effect of several risk factors on survival time. However, these standard survival models encounter some challenges when it comes to real-world datasets.
For instance, they cannot easily capture nonlinear relationships between the covariates. In addition, in many applications, the presence of high-dimensional data is quite common, e.g., gene expression data; however, these traditional methods are not able to efficiently deal with such high-dimensional data. As a result, machine learning-based techniques have become increasingly popular in the survival analysis context over recent years \cite{wang2019machine}. Applying machine learning methods directly to censored data is challenging since the value of a measurement or observation is only partially known. Several studies have successfully modified machine learning algorithms to make use of censored information in survival analysis, e.g., decision trees \cite{gordon1985tree}, artificial neural networks (ANN) \cite{faraggi1995neural}, and support vector machines (SVM) \cite{khan2008support} to name a few. Popular ensemble-based frameworks include bagging survival trees \cite{hothorn2004bagging} and random survival forests \cite{ishwaran2008random}. Also, more advanced learning tasks such as active learning \cite{vinzamuri2014active} and transfer learning \cite{li2016transfer} have been extended toward survival analysis. 

Long-term follow-up of patients is often expensive, both time- and effort-wise and financially. As a result, the number of subjects that are included in a study and followed in time is often limited. However, many more subjects may exist (e.g., through retrospective data collection) that meet the inclusion/exclusion criteria of the follow-up study.
If the study aims to predict outcomes based on variables collected at baseline, then we hypothesize that these extra (unlabeled) data points might actually boost the predictive performance of the resulting model, if used wisely. This corresponds to a semi-supervised learning set-up \cite{chapelle2009semi}, which deals with scenarios where only a small part of the instances in the training data have an outcome label attached, but the rest is unlabeled. To our knowledge, such a semi-supervised learning set-up has never been investigated in the context of survival analysis, and with this article, we aim to fill this gap.

Including unlabeled instances in a survival analysis task leads to three distinct subsets of data, that differ in the amount of supervised information they contain: a set of (1) fully observed, (2) partially observed (censored), and (3) unobserved data points. Our goal is to look at these three subsets of data altogether. In particular, we address two research questions: (1) can the predictive performance over an independent test set be increased by including unlabeled instances (i.e., does the semi-supervised learning setting carry over to the survival analysis context)?, and (2) what is the best approach to integrate the 3 subsets of data in the analysis? 
To address this second question, we propose and compare three different approaches.
The first approach is to treat the unlabeled instances as censored with the censoring time equal to zero and apply a machine learning-based survival analysis technique. 
For the second approach, we apply a standard semi-supervised learning approach. In particular, we use the widely used self-training wrapper technique \cite{yarowsky1995unsupervised,li2005setred}. This technique first builds a classifier over the labeled (in our case, observed and censored) data points and iteratively augments the labeled set with highly confident predictions over the unlabeled dataset.
In the third approach, we propose an adaptation of the second one, in which we initially add the censored instances to the unlabeled set, and exploit the censored information in the data augmentation process, to decide how many instances to add to the labeled set in each iteration.
In all three approaches, we use random survival forests as base learner \cite{ishwaran2008random}. 
In order to answer the research questions, we apply and compare the approaches using fifteen real-life datasets from the healthcare domain.

The remainder of this article is organized as follows. Section \ref{sec:background} introduces the background and reviews some concepts of the employed models including random survival forest and self-training approaches. Section \ref{sec:related work} describes related work. In section \ref{sec:methodology}, three proposed approaches are introduced, two of which are a self-training-based framework that copes with the survival data. Section \ref{sec:Experimental set-up} presents the experimental set-up, including dataset description, unlabeled data generation, performance evaluation, and comparison methods and parameter instantiation. Results are presented in section \ref{sec:Results}. Conclusions are drawn in section \ref{sec:conlcusion }.

\section{Background} \label{sec:background}
In this section, we first review some concepts of using machine learning methods for survival analysis. Afterward, we explain the self-training technique and how one can apply it to a survival analysis problem.

\subsection{Random survival forest}

Random survival forests are well-known ensemble-based learning models that have been widely used in many survival applications and have been shown to be superior to traditional survival models \cite{miao2015random}. Random survival forest (RSF) \cite{ishwaran2008random} is quite close to the original Random Forest by Breiman \cite{breiman2001random}. The random forest algorithm makes a prediction based on tree-structured models. Similar to the random forest, RSF combines bootstrapping, tree building, and prediction aggregating. However, in the splitting criterion to grow a tree and in the predictions returned in the leaf nodes, RSF  explicitly considers survival time and censoring information. 
RSF has three main steps. As the first step, it draws $B$ bootstrap samples from the original data.
In the second step, for each bootstrap sample, a survival tree is grown. At each node of a tree, $p$ candidate variables are randomly selected, where $p$ is a parameter, often defined as a proportion of the original number of variables. The task is to split the node into two child nodes using the best candidate variable and split point, as determined by the log-rank test \cite{segal1988regression}. The best split is the one that maximizes survival differences between the two child nodes. Growing the obtained tree structure is continued until a stop criterion holds (e.g., until the number of observed instances in the terminal nodes drops below a specified value). In the last step, the cumulative hazard function (CHF) associated with each terminal node in a tree is calculated by the Nelson-Aalen estimator, which is a non-parametric estimator of the CHF \cite{kaplan1958nonparametric}.
 All cases within the same terminal node have the same CHF. The ensemble CHF is constructed as the average over the CHF of the $B$ survival trees.  

 Noteworthy, the survival function and  cumulative hazard function as linked as follows \cite{miller1981survival}:
 \[S(t) = e^{-H(t)}\] 
 where $H(t)$ and $S(t)$ denote the cumulative hazard function and the survival function, respectively.

\subsection{Self-training method}
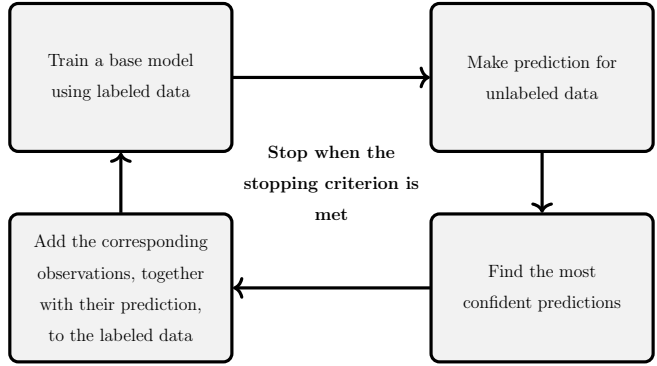
\begin{figure}[!t]
    \centering
 \centering\scalebox{.8}{\begin{tikzpicture}[scale=0.7, transform shape]
\tikzstyle{block} = [rectangle, draw, line width = 1.5pt, minimum height = 10em, minimum width=15em, fill=gray!10, rounded corners]
\tikzstyle{connector} = [->, line width = 1.5pt]

\node[block] (B1) at (0, 0) {
    \begin{minipage}{12em} 
            \begin{center}
               \large Add the corresponding observations, together with their prediction, to the labeled data 
            \end{center}
    \end{minipage}
};
\node[block] (B2) at (0, 5) {
    \begin{minipage}{12em} 
        \begin{center}
            \large Train a base model using labeled data
        \end{center} 
    \end{minipage}
};
\node[block] (B3) at (10, 0) {
    \begin{minipage}{12em} 
             \begin{center}
                \large Find the most confident
            predictions
             \end{center}   
    \end{minipage}
};
\node[block] (B4) at (10, 5) {
    \begin{minipage}{12em} 
             \begin{center}
                \large Make prediction for unlabeled data
             \end{center}  
    \end{minipage}
};

\node (text) at (5, 2.5) {
    \begin{minipage}{12em} 
             \begin{center}
                \large \textbf{Stop when the stopping
                 criterion is met }
             \end{center} 
    \end{minipage}
};

\draw[connector] (B1) -- (B2);
\draw[connector] (B2) -- (B4);
\draw[connector] (B4) -- (B3);
\draw[connector] (B3) -- (B1);

\end{tikzpicture}}
    \caption{Self-training framework. The framework takes  a set of labeled and unlabeled data instances as input and starts in the top left box.}
    \label{fig:Self-training framework}
\end{figure}

The semi-supervised learning (SSL) paradigm is a combination of supervised and unsupervised learning and has been widely used in many applications such as healthcare \cite{madani2018semi,rogers2019bayesian,ballinger2018deepheart}. The primary goal of SSL methods is to take advantage of the unlabeled data in addition to the labeled data, in order to obtain a better prediction model. The acquisition of labeled data is usually expensive, time-consuming, and often difficult, specifically when it comes to healthcare and follow-up data. Hence, achieving good performance with supervised techniques is challenging, since the number of labeled instances is often too small. Over the years, many SSL techniques have been proposed \cite{zhu2005semi,van2020survey}. In this article, we will focus on self-training (sometimes also called self-learning) \cite{yarowsky1995unsupervised}, one of the most widely used algorithms for SSL. Self-training has been used in different approaches like deep neural networks \cite{collobert2008unified}, face recognition \cite{roli2006semi}, and parsing \cite{mcclosky2006effective}.
This framework overcomes the issue of insufficient labeled data by augmenting the training set with unlabeled instances. It starts with training a model using a base learner on the labeled set and then augments this set with the predictions for the unlabeled instances that the model is most confident in (see Figure \ref{fig:Self-training framework}). This procedure is repeated until a certain stopping criterion is met. This stopping criterion, the number of instances to augment in each iteration, and the definition of confidence are instantiated according to the problem at hand.

\section{Related work} \label{sec:related work}

Semi-supervised learning (SSL) methods have been applied in many different domains \cite{zhu2005semi,van2020survey}. 
However, few efforts have been made in order to generalize SSL algorithms to be suitable for survival analysis.

Bair and Tibshirani \cite{bair2004semi} combine supervised and unsupervised learning to predict survival times for cancer patients. They first employ a supervised approach to select a subset of genes from a gene expression dataset that correlates with survival. Then, unsupervised clustering is applied to these gene subsets to identify cancer subtypes. Once such subtypes are identified, they apply again supervised learning techniques to classify future patients into the appropriate subgroup or to predict their survival. Although the authors call the resulting approach semi-supervised, their setting is clearly different from ours.

There has also been some work that models a survival analysis task as a  semi-supervised learning problem by employing a self-training strategy to predict event times from observed and censored data points. Both \cite{shi2011semi, hassanzadeh2016multi} treat the censored data points as unlabeled, thereby ignoring the time-to-event information that they contain. Liang et al \cite{liang2016cancer} do use some information from the censored times, in the sense that they disregard data points for which the model predicts a value lower than the right-censored time points. They combine Cox proportional hazard (Cox) and accelerated failure time (AFT) model in a semi-supervised set-up to predict the treatment risk and the survival time of cancer patients. Regularization is used for gene selection, which is an essential task in cancer survival analysis. The authors found that many censored data points always violate the constraint that the predicted survival time should be higher than the censored time, restricting the full exploitation of the censored data. Therefore, in follow-up work \cite{chai2017new}, they embedded a self-paced learning mechanism in their framework to gradually introduce more complex data samples in the training process, leading to a more accurate estimation for the censored samples. An important diﬀerence between our work and the discussed studies is that we consider situations where apart from fully observed and censored instances,
we also have a third category, namely extra data points that are unlabeled.
To our knowledge, this is the ﬁrst study to investigate the use of unlabeled instances in the survival context.

\section{Methodology} \label{sec:methodology}
In order to predict event times in the presence of observed, censored, and unlabeled instances, we propose three approaches.

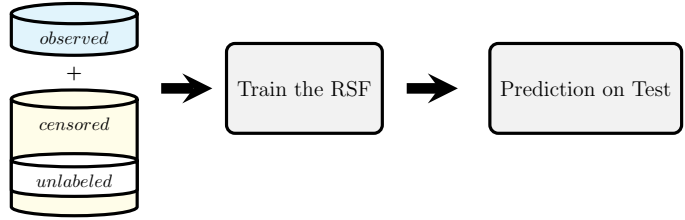
\begin{figure}[!t]
    \centering
    \centering\scalebox{.8}{\begin{tikzpicture}[scale=0.85, transform shape]

    \tikzstyle{block1} = [cylinder, draw, shape border rotate=90, line width = 1.5pt, minimum height = 1.5em, minimum width=7em, shape aspect=.15, fill=white]
    \tikzstyle{block2} = [cylinder, draw, shape border rotate=90, line width = 1.5pt, minimum height = 2.5em, minimum width=7em, shape aspect=.25, fill=cyan!10]
    \tikzstyle{block3} = [cylinder, draw, shape border rotate=90, line width = 1.5pt, minimum height = 7em, minimum width=7em, shape aspect=1.5, fill=yellow!10]
    \tikzstyle{block4} = [rectangle, draw, line width = 1.5pt, minimum height = 5em, minimum width=5em, fill=gray!10, rounded corners]
    \tikzstyle{block5} = [rectangle, draw,dashed, line width = 1pt, minimum height = 15em, minimum width=50em]
    \tikzstyle{block6} = [rectangle, draw,dashed, line width = 1pt, minimum height = 18em, minimum width=50em]
    \tikzstyle{connector} = [->line width = 7pt]
    \tikzstyle{line} = [line width = 1.5pt]
    \tikzstyle{branch} = [circle, inner sep = 0pt, minimum size = 0.5mm, fill = black, draw = black]
    \tikzstyle{arrow} = [>={Triangle[width=6mm,length=6mm]},line width = 5pt,->,>=stealth]


    \node[block2] (B1) at (0, 2.5) {\textbf{$ observed $}};
    \node (X) at (0, 1.8) {\textbf{+}};
    \node[block3] (B2) at (0, 0.1) {};
    \node[block1] (B3) at (0, -.3) {\textbf{$unlabeled$}};
    
    \node (X)  at (0,0.8) {$censored$};
    \draw[arrow] (1.7,1.5) -- (2.7,1.5);
    \node[block4] (B4) at (4.5, 1.5) {
        \begin{minipage}{8em} 
            \begin{center}
                \large Train the RSF
            \end{center}
        \end{minipage}
    };
    \draw[arrow] (6.5,1.5) -- (7.5,1.5);
     \node[block4] (B5) at (10, 1.5) {
        \begin{minipage}{10em} 
            \begin{center}
                \large Prediction on Test
            \end{center}
        \end{minipage}
    };
\end{tikzpicture}}
    \caption{Pipeline for the first approach, called RSF+UD.}
    \label{fig:first approach}
\end{figure}

The first approach is a straightforward application of a survival analysis method (in our case, RSF), in which we add the unlabeled set as censored instances, with the corresponding event time set to zero. We call the first approach random survival forest with unlabeled data (RSF+UD). Figure \ref{fig:first approach} depicts the block diagram of the first proposed pipeline.

\begin{figure}[!t]
    \centering
    \centering\scalebox{.8}{\begin{tikzpicture}[scale=0.85, transform shape]
    
    \tikzstyle{block1} = [cylinder, draw, shape border rotate=90, line width = 1.5pt, minimum height = 3em, minimum width=7em, shape aspect=.25]
    \tikzstyle{block2} = [cylinder, draw, shape border rotate=90, line width = 1.5pt, minimum height = 2.5em, minimum width=7em, shape aspect=.25, fill=cyan!10]
    \tikzstyle{block3} = [cylinder, draw, shape border rotate=90, line width = 1.5pt, minimum height = 2.5em, minimum width=7em, shape aspect=.25, fill=yellow!10]
    \tikzstyle{block4} = [rectangle, draw, line width = 1.5pt, minimum height = 5em, minimum width=5em, fill=gray!10, rounded corners]
    \tikzstyle{block6} = [rectangle, draw,dashed, line width = 1pt, minimum height = 32em, minimum width=28em]
    \tikzstyle{block7} = [circle, draw, line width = 1.5pt,fill=gray!10]
    \tikzstyle{block8} = [cylinder, draw, shape border rotate=90, line width = 1.5pt, minimum height =7em, minimum width=9em, shape aspect=.15, fill=gray!10]
    \tikzstyle{block9} = [diamond, draw, shape border rotate=90, line width = 1.5pt, inner sep=1pt, fill=gray!10]
    \tikzstyle{connector} = [->line width = 7pt]
    \tikzstyle{line} = [line width = 5.5pt]
    \tikzstyle{branch} = [circle, inner sep = 0pt, minimum size = 0.5mm, fill = black, draw = black]
    \tikzstyle{arrow} = [>={Triangle[width=6mm,length=6mm]},line width = 5pt,->,>=stealth]

    \node[block6] (B6) at (7, -7.5) {};
    \node[block2] (B1) at (0, -2.9) {\textbf{$ observed $}};
    \node (X) at (0, -3.5) {\textbf{+}};
    \node[block3] (B2) at (0, -4.4) {\textbf{$ censored $}};
    \node[block1] (B3) at (9.5, -0.7) {\textbf{$ unlabeled $}};
    \draw[arrow] (9.5,-1.2) -- (9.5,-2.1);
    \draw[arrow] (1.7,-3.5) -- (2.7,-3.5);
    \node (F) at (4, -1.5) {\textbf{Self-training approach}};
    \node[block4] (B4) at (4.5, -3.5) {
        \begin{minipage}{8em} 
            \begin{center}
                \large Train the RSF
            \end{center}
        \end{minipage}
    };
    \draw[arrow] (6.5,-3.5) -- (8,-3.5);
    \node[block7] (B5) at (9.5, -3.5) {
        \begin{minipage}{6em} 
            \begin{center}
                \large Trained model
            \end{center}
        \end{minipage}
    };
 \draw[arrow] (9.5,-5.1) -- (9.5,-6.1);
     \node[block8] (B7) at (9.5, -8) {
     \begin{minipage}{9em} 
         \begin{center}
             \large Predicted survival times for the unlabeled observations
         \end{center}
     \end{minipage}
 };
\draw[line] (9.5, -9.5) -- (9.5,-11.5) {};
 \draw[arrow] (9.5,-11.41) -- (6.5,-11.41);
  \node[block9] (B7) at (4.8, -11.41) {
     \begin{minipage}{5em} 
         \begin{center}
             \large Selection criteria
         \end{center}
     \end{minipage}
 };
 \draw[arrow] (4.8,-9.8) -- (4.8,-8.8);
 
   \node[block8] (B7) at (4.8, -7.4) {
     \begin{minipage}{12em} 
         \begin{center}
             \large Selected unlabeled to be added to the labeled observations
         \end{center}
     \end{minipage}
 };
\draw[arrow] (4.8,-5.5) -- (4.8,-4.5);
\draw[arrow] (12,-7.5) -- (13,-7.5);
  \node[block4] (B5) at (15, -7.5) {
    \begin{minipage}{10em} 
        \begin{center}
            \large Prediction on Test
        \end{center}
    \end{minipage}
};

\end{tikzpicture}}
    \caption{Pipeline for the second approach, called ST-RSF.}
    \label{fig:second approach}
\end{figure}
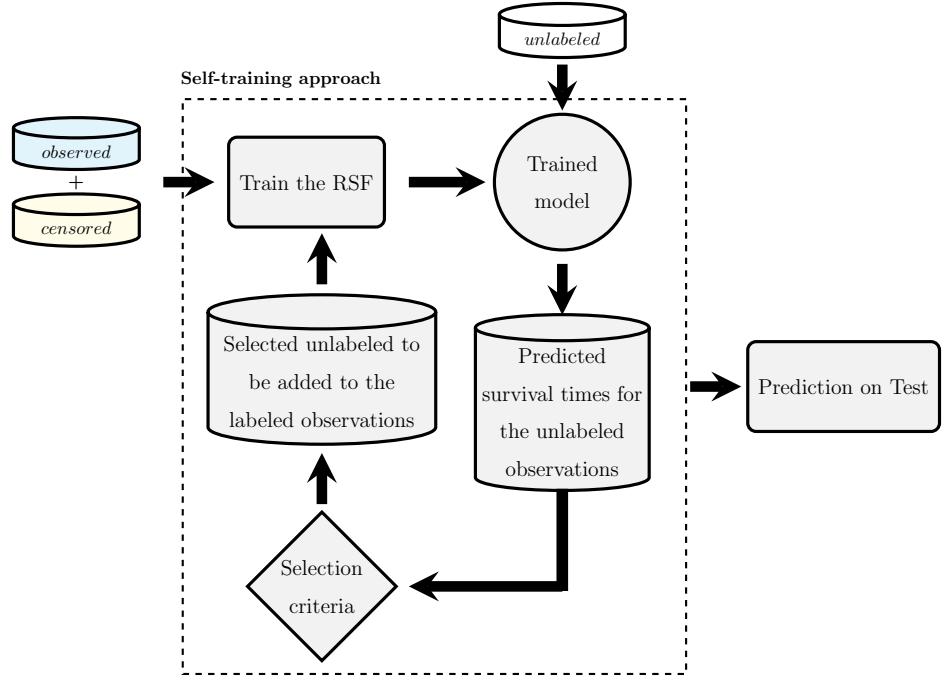

In the second approach, we apply a semi-supervised learning approach called self-trained random survival forest (ST-RSF). In particular, we use the widely used self-training wrapper technique \cite{nigam2000analyzing}. Figure \ref{fig:second approach} shows the learning process in our self-training algorithm. This technique first builds an initial model using RSF over the labeled (in our case, observed and censored) data points and then iteratively augments the labeled set with the most confident predictions of survival time for the unlabeled dataset. In order to predict the survival time for each individual, we calculate the expected future lifetime ($T_p$) which at a given time $t_0$ is the time remaining until the event, given that the event did not occur until $t_0$ \cite{miller1981survival}:
\begin{equation} \label{eq:survival time}
    T_p =\frac{1}{S(t_0)}\int_{t_0}^{\infty} S(t)dt 
\end{equation}
where $S(t)$ is the survival function predicted by RSF.

The aim is to boost the performance of the model using unlabeled data through an iterative process. 
As mentioned in Section~\ref{sec:background}, the adoption of a self-training approach requires the instantiation of three aspects. First, in order to define the confidence in a prediction, we use the variance of predictions across trees. The lower this variance, the more the trees agree, and thus, the more confidence we have in the predicted value. Second, we set the number of instances to be added to the labeled set in each iteration to 10\% of the size of the unlabeled set. We set the status of these newly added instances to observed and add their predicted value as their survival time. Finally, we need to define a global stopping criterion, to terminate the iterative procedure. For this purpose, in the first iteration, we take the first quartile of the variance values and use it as the maximally allowed variance in the whole procedure. Thus, we only augment unlabeled instances if their prediction variance is smaller than this value. If no instances can be added, the algorithm stops.

The details of this approach (ST-RSF) are described in Algorithm \ref{alg: algorithm1}. 

\begin{algorithm}[!b]
    \caption{Self-trained random survival forest (ST-RSF). \label{alg: algorithm1}}    
    \KwIn{labeled data ($Ldata$), unlabeled data ($Udata$)}
    \KwOut{Prediction model for survival time}

   \Repeat{no confident predictions have been added to the training set}{
    Train a base model using $Ldata$\;
    Make a prediction for the survival time ($ T_{p} $) of each instance in  $Udata$ using Equation \ref{eq:survival time}\;
    Calculate the variance for each prediction\;
    Sort the predictions based on minimum variance\;
    If the stopping criterion is not defined yet ($S=-\infty$), find the first quartile as the stopping criterion (only in the first iteration)\;
    Select the top 10\% $Udata$ instances from the sorted list of predictions, with variance smaller than $S$ (confident predictions)\;
    Remove the confident predictions from $Udata$ and add them to the training set ($Ldata$)\; 
    }
\end{algorithm}

The third approach is an adaptation of the second one, where we exploit the information contained in the censored instances to replace the arbitrarily set stopping criterion of the second approach. In particular, we use the self-training wrapper technique as before, but build the initial model over only the observed data points and iteratively augment the training set with high confidence predictions from the censored and unlabeled dataset. In other words, we treat the censored examples as unlabeled, and the observed examples as labeled, and cast the problem as a pure semi-supervised learning problem. However, in this scenario, the censored instances are not totally unlabeled, since we know that their event time is greater than the censoring time (assuming right-censored instances). As a result, we aim to exploit this information of censored instances to introduce a smarter stopping criterion in the data augmentation process. 

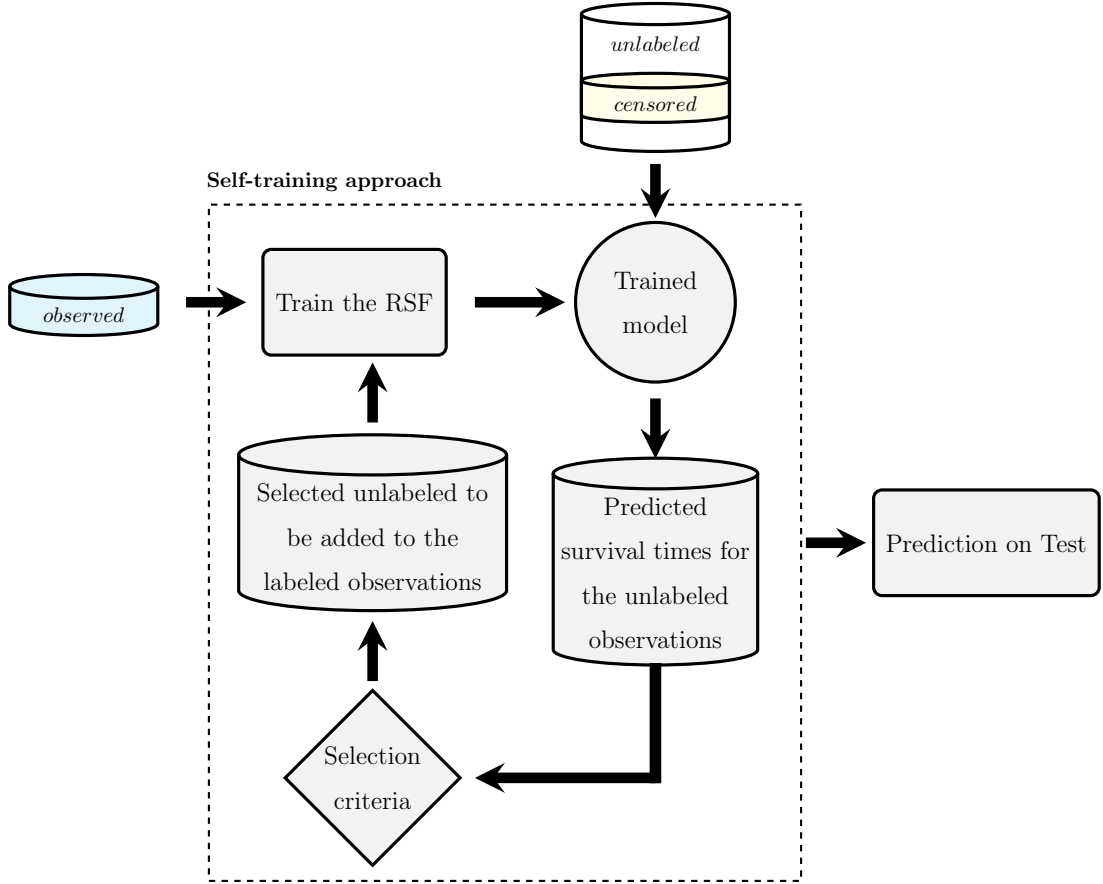
\begin{figure}[!t]
    \centering
    \centering\scalebox{.8}{\begin{tikzpicture}[scale=1, transform shape]
    
    \tikzstyle{block1} = [cylinder, draw, shape border rotate=90, line width = 1.5pt, minimum height = 7em, minimum width=7em, shape aspect=1.5]
    \tikzstyle{block2} = [cylinder, draw, shape border rotate=90, line width = 1.5pt, minimum height = 2.5em, minimum width=7em, shape aspect=.25, fill=cyan!10]
    \tikzstyle{block3} = [cylinder, draw, shape border rotate=90, line width = 1.5pt, minimum height = 1.5em, minimum width=7em, shape aspect=.15, fill=yellow!10]
    \tikzstyle{block4} = [rectangle, draw, line width = 1.5pt, minimum height = 5em, minimum width=5em, fill=gray!10, rounded corners]
    \tikzstyle{block6} = [rectangle, draw,dashed, line width = 1pt, minimum height = 32em, minimum width=28em]
    \tikzstyle{block7} = [circle, draw, line width = 1.5pt,fill=gray!10]
    \tikzstyle{block8} = [cylinder, draw, shape border rotate=90, line width = 1.5pt, minimum height =7em, minimum width=9em, shape aspect=.15, fill=gray!10]
    \tikzstyle{block9} = [diamond, draw, shape border rotate=90, line width = 1.5pt, inner sep=1pt, fill=gray!10]
    \tikzstyle{connector} = [->line width = 7pt]
    \tikzstyle{line} = [line width = 5.5pt]
    \tikzstyle{branch} = [circle, inner sep = 0pt, minimum size = 0.5mm, fill = black, draw = black]
    \tikzstyle{arrow} = [>={Triangle[width=6mm,length=6mm]},line width = 5pt,->,>=stealth]

    \node[block6] (B6) at (7, -7.5) {};
    \node[block2] (B1) at (0, -3.7) {\textbf{$ observed $}};
    \node[block3] (B2) at (9.5, -.2) {\textbf{$ censored $}};
    \node[block1] (B3) at (9.5, 0.1) {\textbf{}};
    \node (X) at (9.5, .8) {$ unlabeled $};
    \draw[arrow] (9.5,-1.2) -- (9.5,-2.1);
    \draw[arrow] (1.7,-3.5) -- (2.7,-3.5);
     \node (F) at (4, -1.5) {\textbf{Self-training approach}};
    \node[block4] (B4) at (4.5, -3.5) {
        \begin{minipage}{8em} 
            \begin{center}
                \large Train the RSF
            \end{center}
        \end{minipage}
    };
    \draw[arrow] (6.5,-3.5) -- (8,-3.5);
    \node[block7] (B5) at (9.5, -3.5) {
        \begin{minipage}{6em} 
            \begin{center}
                \large Trained model
            \end{center}
        \end{minipage}
    };
 \draw[arrow] (9.5,-5.1) -- (9.5,-6.1);
     \node[block8] (B7) at (9.5, -8) {
     \begin{minipage}{9em} 
         \begin{center}
             \large Predicted survival times for the unlabeled observations
         \end{center}
     \end{minipage}
 };
\draw[line] (9.5, -9.5) -- (9.5,-11.5) {};
 \draw[arrow] (9.5,-11.41) -- (6.5,-11.41);
  \node[block9] (B7) at (4.8, -11.41) {
     \begin{minipage}{5em} 
         \begin{center}
             \large Selection criteria
         \end{center}
     \end{minipage}
 };
 \draw[arrow] (4.8,-9.8) -- (4.8,-8.8);
 
   \node[block8] (B7) at (4.8, -7.4) {
     \begin{minipage}{12em} 
         \begin{center}
             \large Selected unlabeled to be added to the labeled observations
         \end{center}
     \end{minipage}
 };
\draw[arrow] (4.8,-5.5) -- (4.8,-4.5);
\draw[arrow] (12,-7.5) -- (13,-7.5);
  \node[block4] (B5) at (15, -7.5) {
    \begin{minipage}{10em} 
        \begin{center}
            \large Prediction on Test
        \end{center}
    \end{minipage}
};

\end{tikzpicture}}
    \caption{Pipeline for the third approach, called ST-RSF+CCT.}
    \label{fig:third approach}
\end{figure}

 We denote this approach as a self-trained random survival forest corrected with censored times (ST-RSF+CCT). Figure \ref{fig:third approach} shows the learning process in this self-training algorithm. 

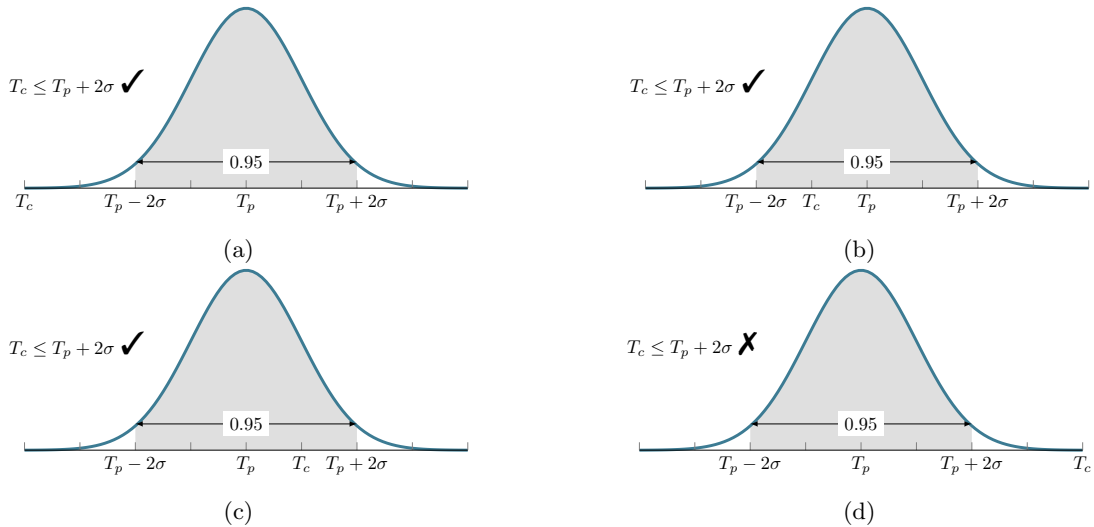
\begin{figure}[!t]
    \centering
    \begin{subfigure}[b]{0.5\textwidth}
     \centering\scalebox{.7}{\pgfmathdeclarefunction{gauss}{2}{%
    \pgfmathparse{1/(#2*sqrt(2*pi))*exp(-((x-#1)^2)/(2*#2^2))}%
}

    \begin{tikzpicture}[scale=1, transform shape]
        \begin{axis}[
            no markers,
            domain=-6:6,
            samples=100,
            axis lines*=left,
            axis y line=none,
            xlabel=,
            ylabel=,
            every axis x label/.style={at=(current axis.right of origin),anchor=west},
            height=5cm, width=10cm,
            enlargelimits=false, clip=false, axis on top,
            xtick={-6,-4.5,...,6},
            xticklabels={$T_{c}$,,$T_{p}-2\sigma$,,$T_{p}$,,$T_{p}+2\sigma$,,}]

            \addplot [  draw=none, domain=-6:-4.5] {gauss(0,1.5)} \closedcycle;
            \addplot [ draw=none, domain=-4.5:-3]   {gauss(0,1.5)} \closedcycle;
            \addplot [ draw=none, domain=-3:-1.5] {gauss(0,1.5)} \closedcycle;
            \addplot [fill=gray,nearly transparent, draw=none, domain=-3: 3] {gauss(0,1.5)} \closedcycle;
            \addplot [draw=none, domain=3:1.5] {gauss(0,1.5)} \closedcycle;
            \addplot [ draw=none, domain=4.5:3]   {gauss(0,1.5)} \closedcycle;
            \addplot [  draw=none, domain=6:4.5] {gauss(0,1.5)} \closedcycle;

            \addplot [ultra thick,cyan!50!black, domain=-6:6] {gauss(0,1.5)};
            
       
            \draw [yshift=0.5 cm, latex-latex](axis cs:-3, 0) -- node [fill=white] {$0.95$} (axis cs:3, 0);
        \end{axis}
        \node (X) at (1,2) {$T_c \leq T_p+2\sigma$ \vspace{2pt}\LARGE\cmark};
    \end{tikzpicture}}
    \caption{}
    \label{fig:cinterval1}
    \end{subfigure}%
    \begin{subfigure}[b]{0.5\textwidth}
          \centering\scalebox{.7}{\pgfmathdeclarefunction{gauss}{2}{%
    \pgfmathparse{1/(#2*sqrt(2*pi))*exp(-((x-#1)^2)/(2*#2^2))}%
}

    \begin{tikzpicture}[scale=1, transform shape]
        \begin{axis}[
            no markers,
            domain=-6:6,
            samples=100,
            axis lines*=left,
            axis y line=none,
            xlabel=,
            ylabel=,
            every axis x label/.style={at=(current axis.right of origin),anchor=west},
            height=5cm, width=10cm,
            enlargelimits=false, clip=false, axis on top,
            xtick={-6,-4.5,...,6},
            xticklabels={,,$T_{p}-2\sigma$,$T_{c}$,$T_{p}$,,$T_{p}+2\sigma$,,}]

            \addplot [  draw=none, domain=-6:-4.5] {gauss(0,1.5)} \closedcycle;
            \addplot [ draw=none, domain=-4.5:-3]   {gauss(0,1.5)} \closedcycle;
            \addplot [ draw=none, domain=-3:-1.5] {gauss(0,1.5)} \closedcycle;
            \addplot [fill=gray,nearly transparent, draw=none, domain=-3: 3] {gauss(0,1.5)} \closedcycle;
            \addplot [draw=none, domain=3:1.5] {gauss(0,1.5)} \closedcycle;
            \addplot [ draw=none, domain=4.5:3]   {gauss(0,1.5)} \closedcycle;
            \addplot [  draw=none, domain=6:4.5] {gauss(0,1.5)} \closedcycle;

            \addplot [ultra thick,cyan!50!black, domain=-6:6] {gauss(0,1.5)};
            
       
            \draw [yshift=0.5 cm, latex-latex](axis cs:-3, 0) -- node [fill=white] {$0.95$} (axis cs:3, 0);
        \end{axis}
         \node (X) at (1,2) {$T_c \leq T_p+2\sigma$ \vspace{2pt} \LARGE\cmark};
    \end{tikzpicture}}
    \caption{}
    \label{fig:cinterval2}
    \end{subfigure}

     \begin{subfigure}[b]{0.5\textwidth}
          \centering\scalebox{.7}{\pgfmathdeclarefunction{gauss}{2}{%
    \pgfmathparse{1/(#2*sqrt(2*pi))*exp(-((x-#1)^2)/(2*#2^2))}%
}

    \begin{tikzpicture}[scale=1, transform shape]
        \begin{axis}[
            no markers,
            domain=-6:6,
            samples=100,
            axis lines*=left,
            axis y line=none,
            xlabel=,
            ylabel=,
            every axis x label/.style={at=(current axis.right of origin),anchor=west},
            height=5cm, width=10cm,
            enlargelimits=false, clip=false, axis on top,
            xtick={-6,-4.5,...,6},
            xticklabels={,,$T_{p}-2\sigma$,,$T_{p}$,$T_{c}$,$T_{p}+2\sigma$,,}]

            \addplot [  draw=none, domain=-6:-4.5] {gauss(0,1.5)} \closedcycle;
            \addplot [ draw=none, domain=-4.5:-3]   {gauss(0,1.5)} \closedcycle;
            \addplot [ draw=none, domain=-3:-1.5] {gauss(0,1.5)} \closedcycle;
            \addplot [fill=gray,nearly transparent, draw=none, domain=-3: 3] {gauss(0,1.5)} \closedcycle;
            \addplot [draw=none, domain=3:1.5] {gauss(0,1.5)} \closedcycle;
            \addplot [ draw=none, domain=4.5:3]   {gauss(0,1.5)} \closedcycle;
            \addplot [  draw=none, domain=6:4.5] {gauss(0,1.5)} \closedcycle;

            \addplot [ultra thick,cyan!50!black, domain=-6:6] {gauss(0,1.5)};
            
       
            \draw [yshift=0.5 cm, latex-latex](axis cs:-3, 0) -- node [fill=white] {$0.95$} (axis cs:3, 0);
        \end{axis}
        \node(X) at (1,2) {$T_c \leq T_p+2\sigma$ \vspace{2pt}\LARGE\cmark};
    \end{tikzpicture}}
    \caption{}
    \label{fig:cinterval3}
    \end{subfigure}%
     \begin{subfigure}[b]{0.5\textwidth}
          \centering\scalebox{.7}{\pgfmathdeclarefunction{gauss}{2}{%
    \pgfmathparse{1/(#2*sqrt(2*pi))*exp(-((x-#1)^2)/(2*#2^2))}%
}

    \begin{tikzpicture}[scale=1, transform shape]
        \begin{axis}[
            no markers,
            domain=-6:6,
            samples=100,
            axis lines*=left,
            axis y line=none,
            xlabel=,
            ylabel=,
            every axis x label/.style={at=(current axis.right of origin),anchor=west},
            height=5cm, width=10cm,
            enlargelimits=false, clip=false, axis on top,
            xtick={-6,-4.5,...,6},
            xticklabels={,,$T_{p}-2\sigma$,,$T_{p}$,,$T_{p}+2\sigma$,,$T_{c}$}]

            \addplot [  draw=none, domain=-6:-4.5] {gauss(0,1.5)} \closedcycle;
            \addplot [ draw=none, domain=-4.5:-3]   {gauss(0,1.5)} \closedcycle;
            \addplot [ draw=none, domain=-3:-1.5] {gauss(0,1.5)} \closedcycle;
            \addplot [fill=gray,nearly transparent, draw=none, domain=-3: 3] {gauss(0,1.5)} \closedcycle;
            \addplot [draw=none, domain=3:1.5] {gauss(0,1.5)} \closedcycle;
            \addplot [ draw=none, domain=4.5:3]   {gauss(0,1.5)} \closedcycle;
            \addplot [  draw=none, domain=6:4.5] {gauss(0,1.5)} \closedcycle;

            \addplot [ultra thick,cyan!50!black, domain=-6:6] {gauss(0,1.5)};
            
       
            \draw [yshift=0.5 cm, latex-latex](axis cs:-3, 0) -- node [fill=white] {$0.95$} (axis cs:3, 0);

        \end{axis}
         \node(X) at (1,2) {$T_c \leq T_p+2\sigma$\vspace{4pt} \LARGE\xmark};
    \end{tikzpicture}}
    \caption{}
    \label{fig:cinterval4}
    \end{subfigure}
            \caption{Tolerance interval corresponding to two times the standard deviation. Figures a, b, and c represent situations where the condition $T_{c}\leq T_{p}+2\sigma$ is fulfilled, where $\sigma$ is the standard deviation of the individual tree predictions, and hence, these situations are accepted by our method. In Figure d, the condition is violated.}
        \label{fig:cinterval}
\end{figure}

When deciding which unlabeled (including censored) instances to add to the augmentation process, similarly to the previous approach, we assess the confidence of the ensemble predictions based on the variance of the individual tree predictions. We sort the predictions based on minimum variance (note that the resulting list contains instances both from the censored and unlabeled dataset), but instead of picking the top 10\%, we use the information in the censored instances to decide when to stop adding instances. More precisely, we know that the true event time must be greater than the censoring time for those instances. As a result, whenever we encounter a censored instance with a predicted time $T_p$ smaller than its censoring time $T_c$, we stop the augmentation for the current iteration. When an iteration yields zero augmented instances, the whole procedure is terminated. Preliminary experiments showed that the condition $T_c \leq T_p$ is often too strict and results in premature termination. This happens when the prediction variances are high, and thus adding or removing some trees from the forest could result in a substantially different $T_p$ value and thus a different condition outcome. For this reason, we calculate the $95\%$ tolerance interval around $T_p$ and require $T_c$ to be smaller than or inside the tolerance  interval. In other words, we allow $T_c$ to be larger than $T_p$, but only if it is within its 95\% tolerance interval (see Figure \ref{fig:cinterval}). For the instances (censored or unlabeled) that meet the criterion to be added to the training set, we set the status to observed with the survival time equal to $T_{p}$. Note that this removes the need to use a machine learning method that is able to work with censored instances as the base learner. In this article, in order to be consistent with and provide a fair comparison to the previous approaches, we still use RSF in this approach. The details of this approach are described in Algorithm \ref{alg: algorithm2}.

\begin{algorithm}[!t]
    \caption{Self-trained random survival forest corrected with censored 
times (ST-RSF+CCT). \label{alg: algorithm2}}    
    \KwIn{observed data ($observed$), censored data ($censored$), unlabeled data ($Udata$)}
    \KwOut{Prediction model for survival time}
   \Repeat{no confident predictions have been added to the training set}{
    Train a base model using $observed$\;
    Make a prediction for the survival time ($ T_{p} $) of each instance in $censored \cup Udata$\ using Equation \ref{eq:survival time}\;
    Calculate the variance for each prediction\;
    Sort the predictions based on minimum variance\;
    Calculate 95\%  tolerance interval corresponding to two times the standard deviation of the individual tree predictions ($T _{p}\pm2\sigma$)  for the instances from $censored$ \;
    Find the first $censored$ instance $i$ from the sorted predictions whose censoring time ($T_{c}$) is greater than $T _{p}+2\sigma$ (does not meet the criterion)\;
    Remove all instances sorted before $i$ (confident predictions) from $censored \cup Udata$ and add them to the training set ($observed$)\;
    }
\end{algorithm}

\section{Experimental set-up} \label{sec:Experimental set-up}
In this section, first, we start with a dataset description and then explain the process of creating an unlabeled dataset. In Section \ref{sec: performance evaluation}, we discuss the metric of evaluation, and finally, in Section \ref{sec:Parameter instantiation }, we explain the comparison methods and parameter instantiation.

\subsection{Dataset description}
 We investigate the performance of our proposed approaches on real-life datasets from the \textit{survival} package \cite{survival-package} in R as well as high-dimensional datasets from \cite{Doe:2009:Online,NHANESI}, and some from the R/Bioconductor package. To assess the effectiveness of the proposed approaches in high-dimensional scenarios  ($ p \gg n $),  we used ten different gene expression datasets. These datasets typically contain the expression levels of thousands of genes across a small number of samples ($ < 300 $), giving information about demographic features, disease type, survival time, etc. For convenience, in datasets with more than 10000 gene expression features, we reduced the total number of features to the top 10000 features with the largest variance across all samples.  Table \ref{table:datasets} shows the description and characteristics of the used datasets.  The prediction task for all datasets is survival time (time to death).

\begin{table}[!t]
    \caption{Characteristics of the used clinical and high-dimensional datasets.} 
    \centering 
    \begin{adjustbox}{width=0.9\columnwidth,center}
        \def\arraystretch{1.4}
        \begin{tabular}{l c c c} 
            \hline\hline
            Name & \#Observations & \#Features & Censoring rate \\ [0.5ex] 
            
            \hline 
            Veteran & 137 & 6 & $6\%$ \\ 
            Lung & 228 & 8 & $27\%$ \\
            PBC & 312 & 17 & $60\%$ \\
            DrAsGiven  & 119 & 22122 & $42\%$ \\
            EMTAB386 & 129 & 10364 & $44\%$ \\
            GSE14764 & 80 & 13112 & $74\%$ \\
            GSE32062 & 260 & 20112 & $54\%$ \\  
            Norway/Stanford Breast Cancer Data (NSBCD) & 115 & 549 & $67\%$ \\ 
            Sporadic lymph-node-negative patients (Veer) & 78 & 4751 & $56\%$ \\
            Dutch Breast Cancer Data (DBCD) & 295 & 4919 & $73\%$ \\
           Diffuse Large-B-Cell Lymphoma data (DLBCL) & 240 & 7399 & $42\%$ \\
            Lung adenocarcinomas (LungBeer) & 86 & 7129 & $72\%$ \\
            Acute myeloid leukemia (AML) & 79 & 54675 & $40\%$ \\
           Breast invasive carcinoma (BRCA) & 1080 & 117 & $86\%$ \\
           First National Health and Nutrition Examination Survey (NHANES I) & 9549 & 21 & $64\%$ \\[1ex] 
            \hline 
        \end{tabular}
    \end{adjustbox}
    \label{table:datasets}
\end{table}

\subsection{Unlabeled data generation}\label{unlabeled making}
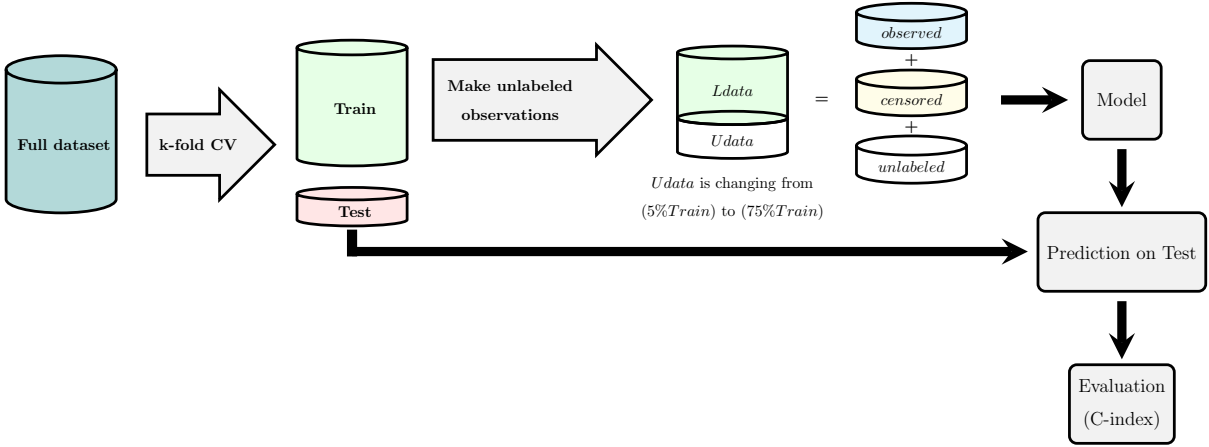
\begin{figure}[t]
    \centering
       \centering\scalebox{.7}{\begin{tikzpicture}[scale=0.85, transform shape]
    \tikzstyle{block1} = [cylinder, draw, shape border rotate=90, line width = 1.5pt, minimum height = 10em, minimum width=7em, shape aspect=.25,fill=teal!30]
    \tikzstyle{block2} = [single arrow, draw, line width = 1.5pt, minimum width=7em, single arrow head extend=.6em,fill=gray!10]
    \tikzstyle{block3} = [cylinder, draw, shape border rotate=90, line width = 1.5pt, minimum height = 8em, minimum width=7em, shape aspect=.25,fill=green!10]
    \tikzstyle{block4} = [cylinder, draw, shape border rotate=90, line width = 1.5pt, minimum height = 2em, minimum width=7em, shape aspect=.25,fill=red!10]
    \tikzstyle{block5} = [cylinder, draw, shape border rotate=90, line width = 1.5pt, minimum height = 5em, minimum width=7em, shape aspect=.25,fill=green!10]
    \tikzstyle{block6} = [cylinder, draw, shape border rotate=90, line width = 1.5pt, minimum height = 3em, minimum width=7em, shape aspect=.25]
    \tikzstyle{block7} = [cylinder, draw, shape border rotate=90, line width = 1.5pt, minimum height = 2.5em, minimum width=7em, shape aspect=.25, fill=cyan!10]
    \tikzstyle{block8} = [cylinder, draw, shape border rotate=90, line width = 1.5pt, minimum height = 2.5em, minimum width=7em, shape aspect=.25, fill=yellow!10]
    \tikzstyle{connector} = [->, line width = 1.5pt]
    \tikzstyle{line} = [line width = 1.5pt]
    \tikzstyle{branch} = [circle, inner sep = 0pt, minimum size = 0.5mm, fill = black, draw = black]
    \tikzstyle{block9} = [rectangle, draw, line width = 1.5pt, minimum height = 5em, minimum width=5em, fill=gray!10, rounded corners]
    \tikzstyle{line} = [line width = 5.5pt]
    \tikzstyle{arrow} = [>={Triangle[width=6mm,length=6mm]},line width = 5pt,->,>=stealth]


    \node[block1] (B1) at (0, 0) {\textbf{ Full dataset }};
    \node[block2] (B2) at (3, 0) {\textbf{ k-fold CV }};
    \node[block3] (B3) at (6.5, 0.8) {\textbf{ Train }};
    \node[block4] (B4) at (6.5, -1.5) {\textbf{ Test }};
    \node[block2] (B5) at (10.3, 1) {
    \begin{minipage}{9em} 
        \begin{center}
            \textbf{Make unlabeled observations}
        \end{center} 
    \end{minipage}\hspace*{15pt}};
    \node[block5] (B6) at (15, 1.21) {\textbf{$ Ldata $}};
    \node[block6] (B7) at (15, 0.1) {\textbf{$ Udata $}};
    \node (Y) at (15, -1.2) {
        \begin{minipage}{12em} 
            \begin{center}
               $Udata$ is changing from ($5\%Train$) to ($75\%Train$)
            \end{center} 
        \end{minipage}
      };
   \node (Z) at (17, 1) {\textbf{=}};
   \node[block7] (B8) at (19, 2.5) {\textbf{$ observed $}};
   \node (X) at (19, 1.9) {\textbf{+}};
   \node[block8] (B8) at (19, 1) {\textbf{$ censored $}};
   \node (Q) at (19, 0.4) {\textbf{+}};
   \node[block6] (B9) at (19, -.5) {\textbf{$ unlabeled $}};
   \draw[arrow] (21,1) -- (22.5,1);
   \node[block9] (B10) at (23.7, 1) {
    \begin{minipage}{4em} 
        \begin{center}
            \large Model
        \end{center}
    \end{minipage}};
    \draw[line] (6.5, -1.9) -- (6.5,-2.5) {};
\draw[arrow] (6.5, -2.39) -- (21.6,-2.39) {};
 \draw[arrow] (23.7,-.1) -- (23.7,-1.4);
  \node[block9] (B11) at (23.7, -2.4) {
    \begin{minipage}{10em} 
        \begin{center}
            \large Prediction on Test
        \end{center}
    \end{minipage}
};
\draw[arrow] (23.7, -3.5) -- (23.7,-4.8) {};
   \node[block9] (B10) at (23.7, -5.8) {
    \begin{minipage}{6em} 
        \begin{center}
            \large Evaluation (C-index)
        \end{center}
    \end{minipage}};

    
\end{tikzpicture}}
    \caption{Illustration of the used procedure in the paper. The first part illustrates the process of making an unlabeled set. Then, the box Model uses one of the three proposed approaches. Predictions are made for the Test set, and finally, evaluations are made using the evaluation metric (C-index).}
    \label{fig:unlabel making}
\end{figure}

 Since we are not aware of survival datasets that include unlabeled instances,  we artificially remove the label of a subset of instances as follows (see Figure \ref{fig:unlabel making}). First, we take the original data and construct five folds for cross-validation, in order to have a fair evaluation of our approach.
Then, for each training set in the cross-validation (i.e., for each combination of four folds), we construct the unlabeled category. We split the training data into two sets called labeled data (Ldata) and unlabeled data (Udata). To have a fair and accurate evaluation, we make sure to have the same distribution for both sets relative to the status (being censored or observed). Then, we take Udata and make the instances unlabeled by removing their time and status values.

To improve the stability of the results, we repeat the cross-validation process 10 times and report the average results. We also vary the percentage of unlabeled instances from 5\% to 75\% of the original training set. 

\subsection{Performance evaluation}	\label{sec: performance evaluation}
In survival analysis, instead of measuring the absolute survival time for each instance, a popular way to assess a model is to estimate the relative risk of an event occurring for different instances. The Harrell's concordance index (C-index) \cite{harrell1982evaluating} is a common way to evaluate a model in survival analysis \cite{schmid2016use}. C-index can be interpreted as the fraction of all pairs of subjects whose predicted survival times are correctly ordered among all subjects that can actually be ordered. In other words, it is the probability of concordance between the predicted and the observed survival time. Two subjects’ survival times can be ordered not only if (1) both of them are observed but also if (2) the observed time of one is smaller than the censored survival time of the other \cite{steck2008ranking}. Consider a set of observation and prediction values for two different instances, $ (y_{1},\hat{y} _{1}) $ and $ (y_{2},\hat{y} _{2}) $, where $ y_{i} $ and $ \hat{y} _{i} $ represent the actual survival time and the predicted value, respectively. The
concordance probability between these two instances can be computed as $ c= Pr(\hat{y} _{1}> \hat{y} _{2}|y_{1}>y_{2}) $. In this paper, we compute the C-index for each test fold in the cross-validation process and return the average value over the 5 test folds. 

\subsection{Comparison methods and parameter instantiation}	\label{sec:Parameter instantiation }
 We applied five different methods: the three methods presented in this article, namely RSF+UD, ST-RSF, ST-RSF+CCT, and  standard RSF and Lasso-Cox trained on the Ldata set only. The goal to perform RSF was to address the first research question (see Section \ref{sec:intro}), i.e., to investigate if adding an unlabeled set to the training phase would increase the performance of the model. The comparison of the three proposed approaches addresses the second research question. To avoid falling into a slightly biased random survival forest comparison, we have reported results of Cox regression with LASSO regularization (Lasso-Cox) as a baseline model. Lasso-Cox  introduces the $L1$ norm penalty in the Cox log-likelihood function \cite{tibshirani1997lasso}. Since the majority of our used datasets are high-dimensional ($ p \gg n $), we have employed Lasso-Cox due to its capability of handling high-dimensional datasets.

In order to estimate the generalization capacity of the models, we performed a 5-fold cross-validation on each dataset and estimate test accuracy, and repeated it 10 times to achieve reliable results.  It is worthwhile to mention that the optimal tuning parameter ($\lambda$) in Lasso-Cox is chosen by nested cross-validation while no hyperparameter tuning has been employed for the other approaches. For RSF-based methods, the number of trees was set to 500, and the number of candidate variables considered in each tree node was set to $p/3$, where $p$ is the number of variables. 

\section{Results and discussion} \label{sec:Results}
\begin{figure}[!t]
    \centering
    \begin{subfigure}{0.32\textwidth}
     \centering
     \includegraphics[width=\linewidth]{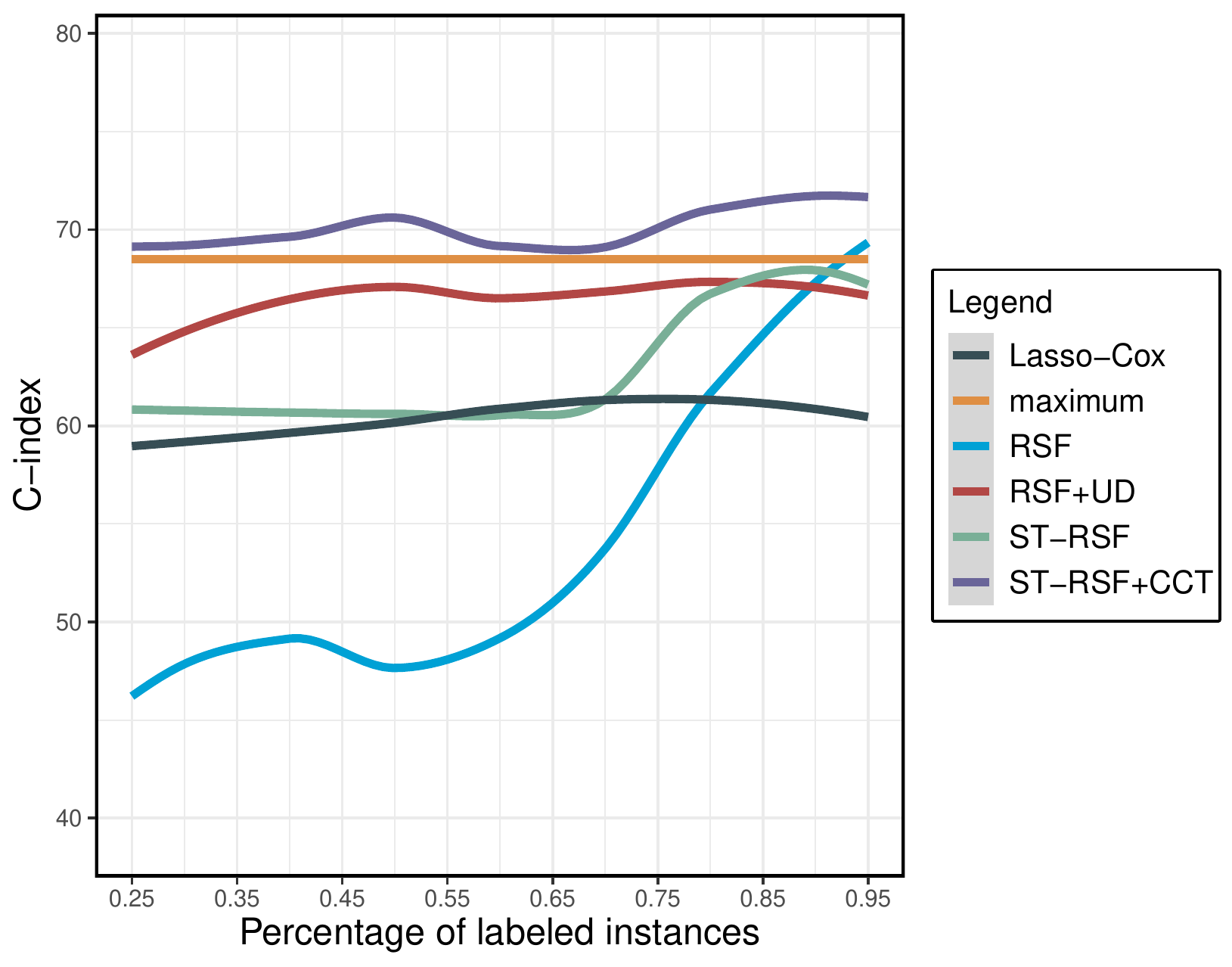}
    \caption{Veer}
    \label{fig:Veer}
    \end{subfigure}%
    \begin{subfigure}{0.32\textwidth}
          \centering
          \includegraphics[width=\linewidth]{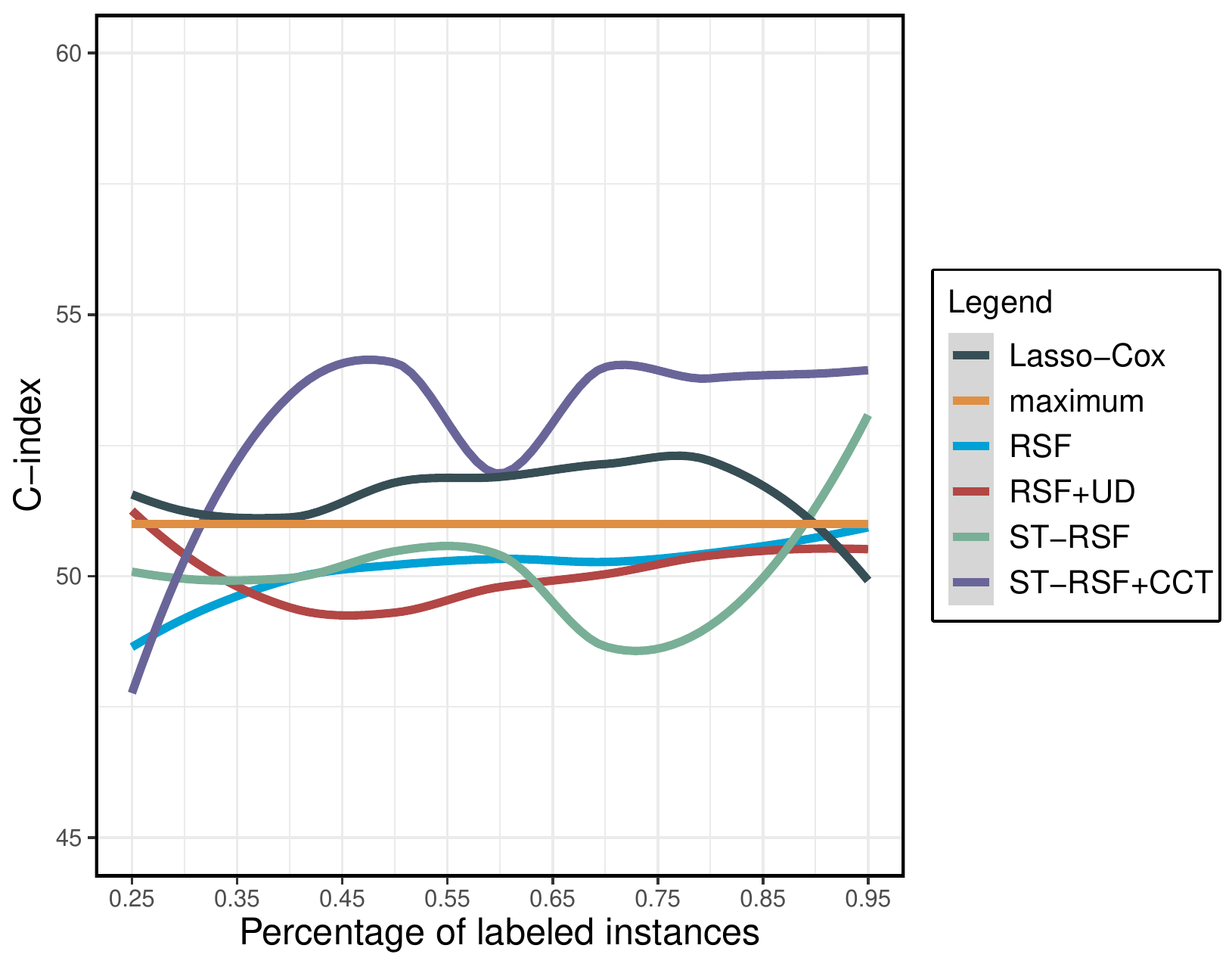}
    \caption{EMTAB386}
    \label{fig:EMTAB}
    \end{subfigure}%
        \begin{subfigure}{0.32\textwidth}
          \centering
          \includegraphics[width=\linewidth]{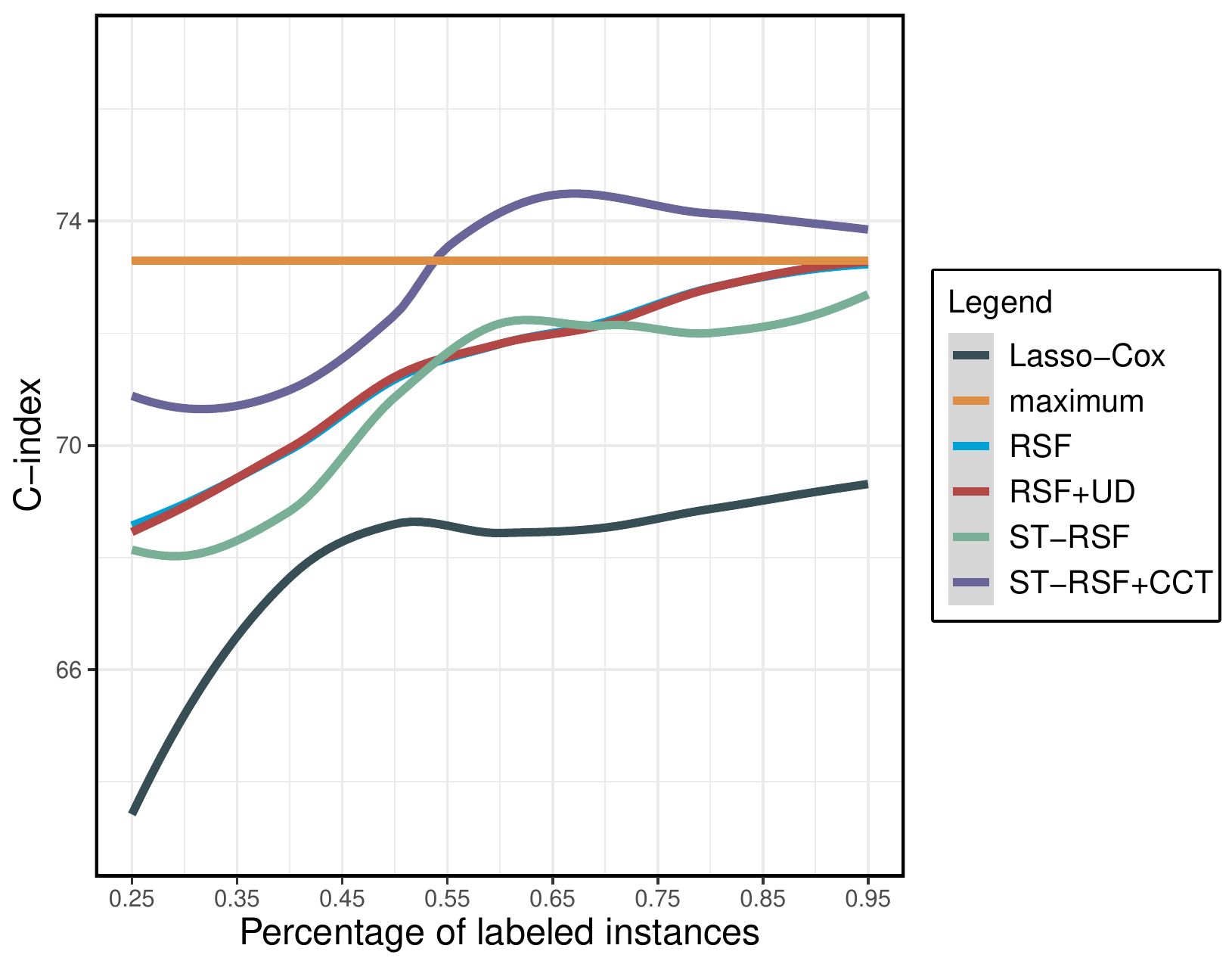}
    \caption{DBCD}
    \label{fig:DBCD}
    \end{subfigure}

     \begin{subfigure}{0.32\textwidth}
          \centering
          \includegraphics[width=\linewidth]{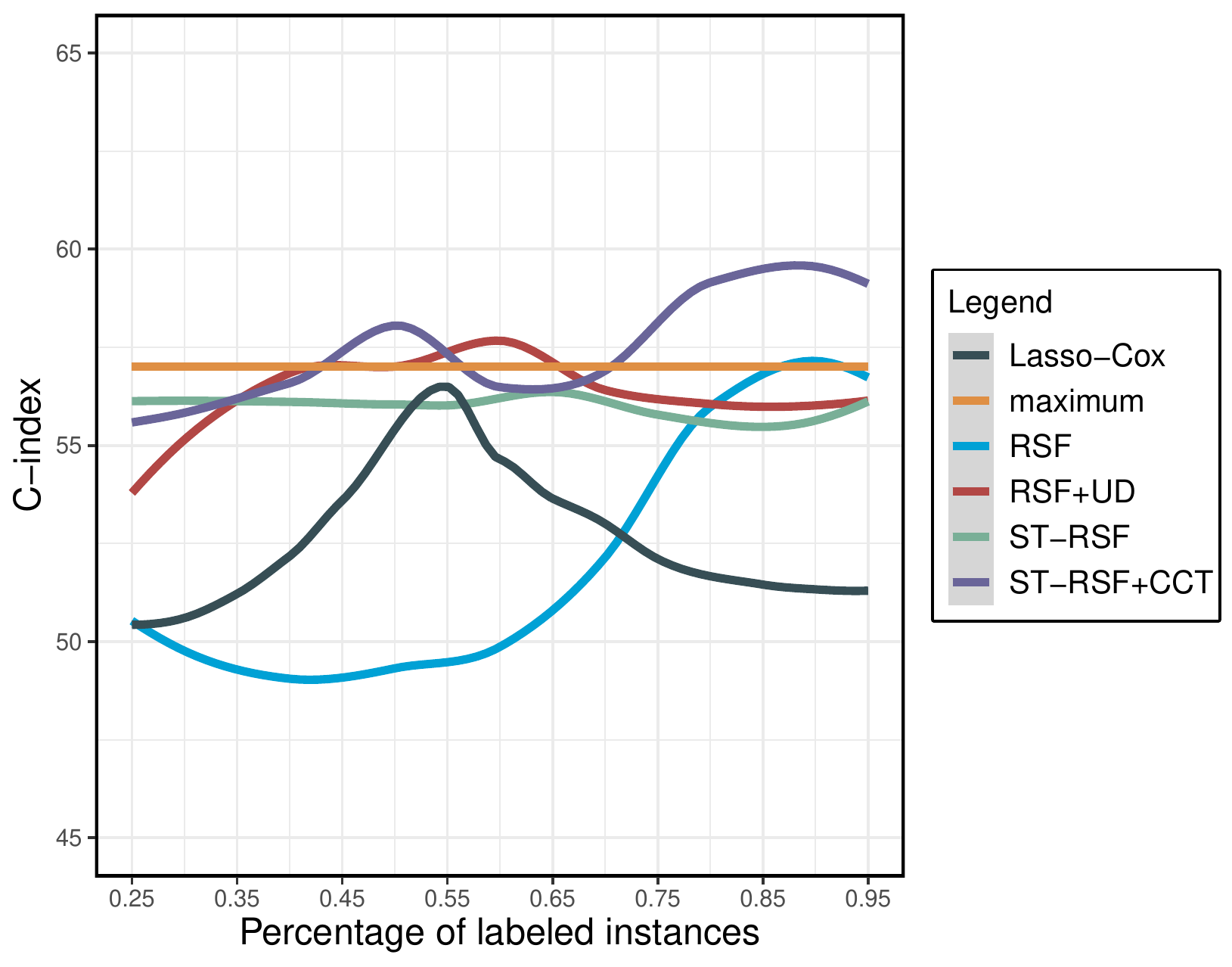}
    \caption{GSE14764}
    \label{fig:GSE14764}
    \end{subfigure}%
     \begin{subfigure}{0.32\textwidth}
          \centering
          \includegraphics[width=\linewidth]{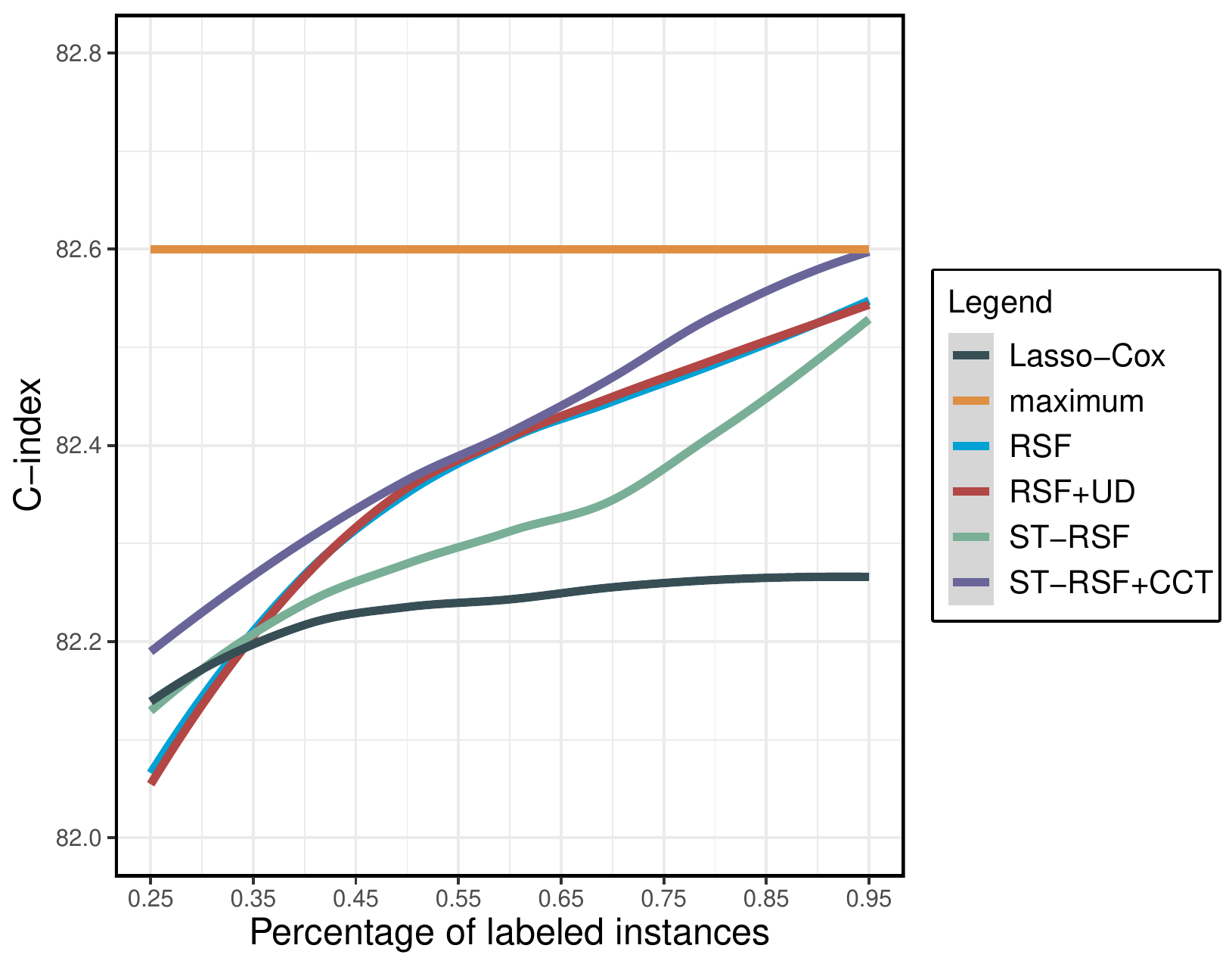}
    \caption{NHANES I}
    \end{subfigure}%
          \begin{subfigure}{0.32\textwidth}
          \centering
          \includegraphics[width=\linewidth]{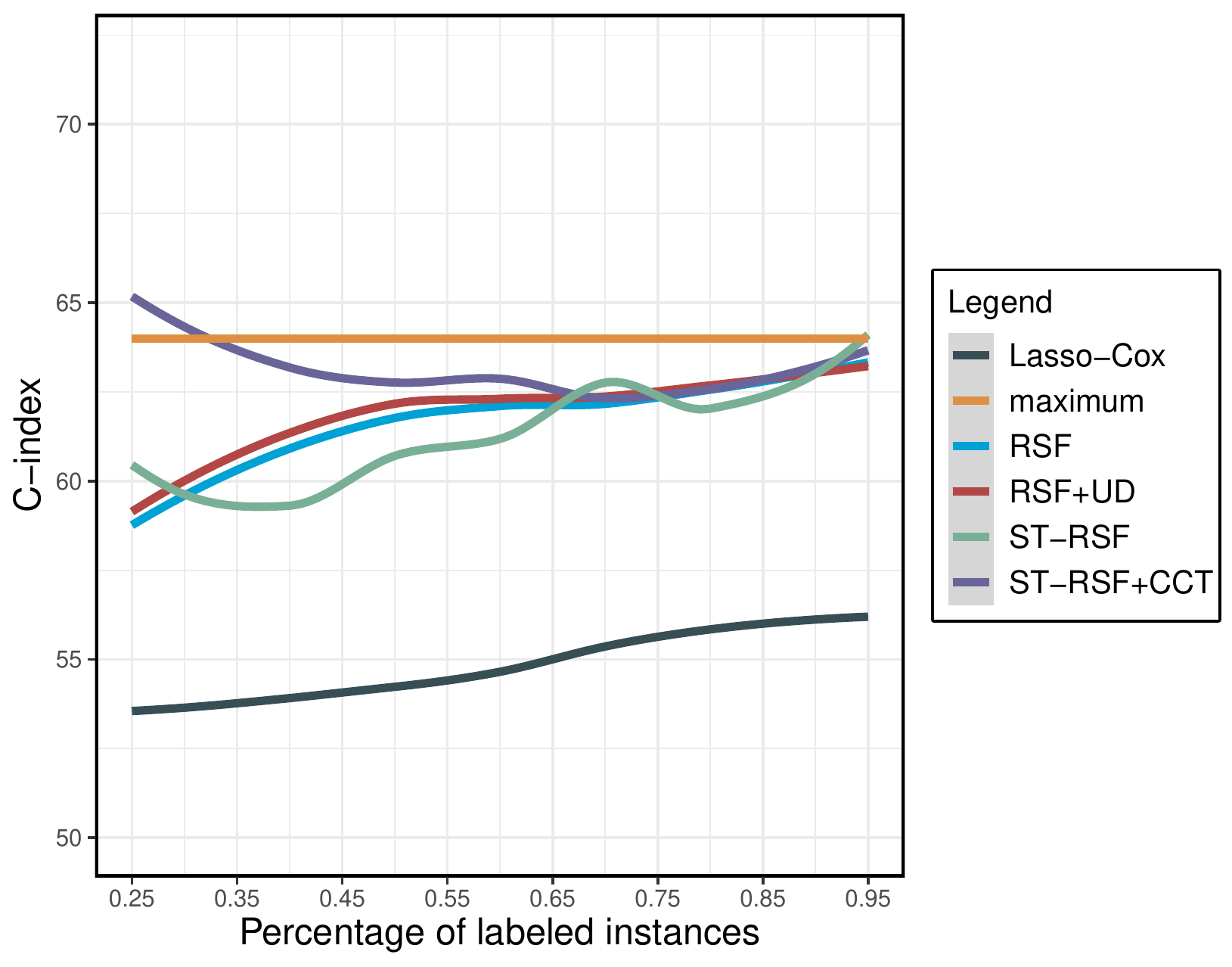}
    \caption{BRCA}
    \end{subfigure}

            \caption{Evaluation of the  performance of the methods, for diﬀerent percentages of labeled instances for six datasets with a high percentage of censored instances.}
        \label{fig:group1}
\end{figure}

\begin{figure}[!t]
    \centering
    \begin{subfigure}{0.32\textwidth}
     \centering
     \includegraphics[width=\linewidth]{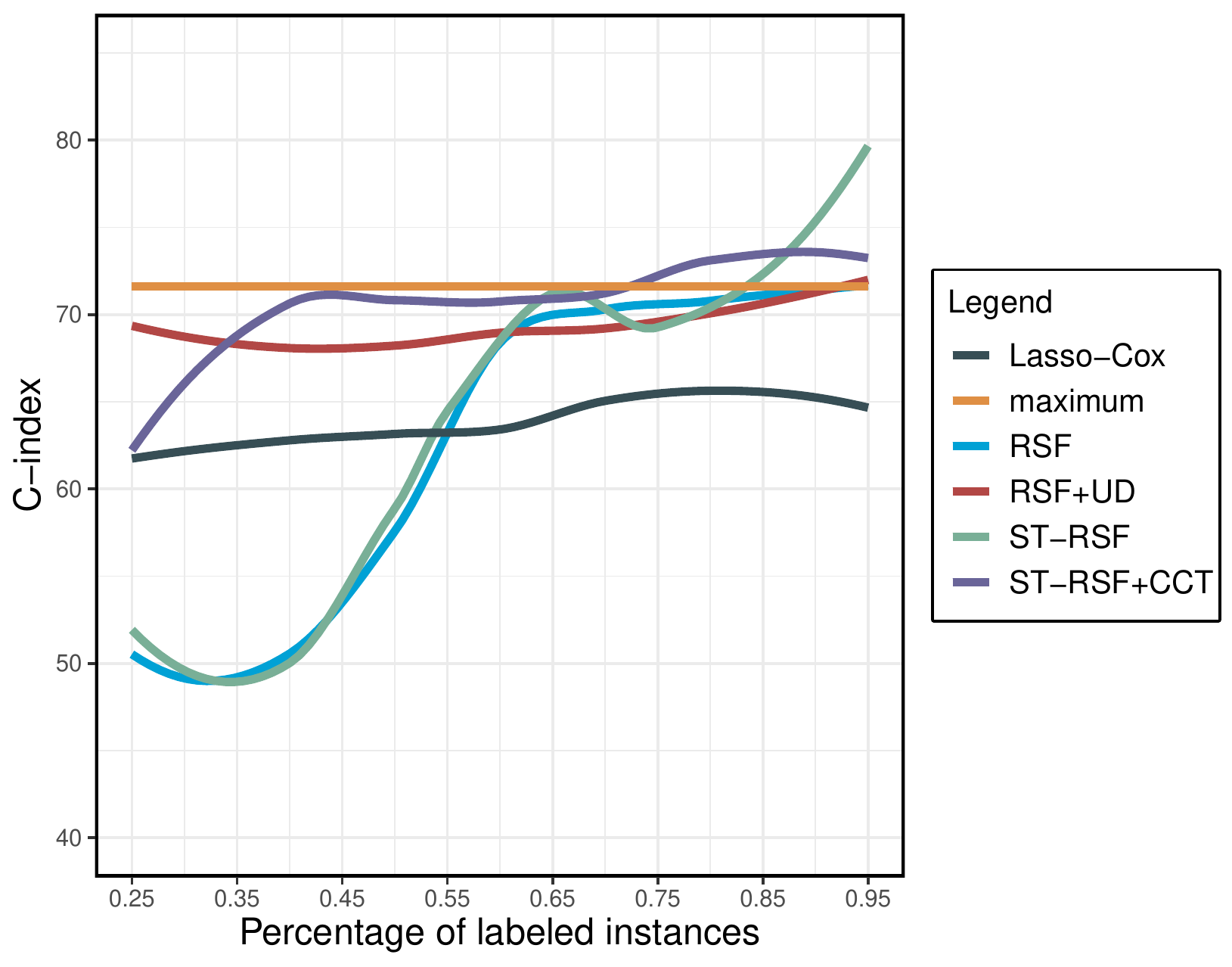}
    \caption{NSBCD}
    \label{fig:NSBCD}
    \end{subfigure}%
    \begin{subfigure}{0.32\textwidth}
          \centering
          \includegraphics[width=\linewidth]{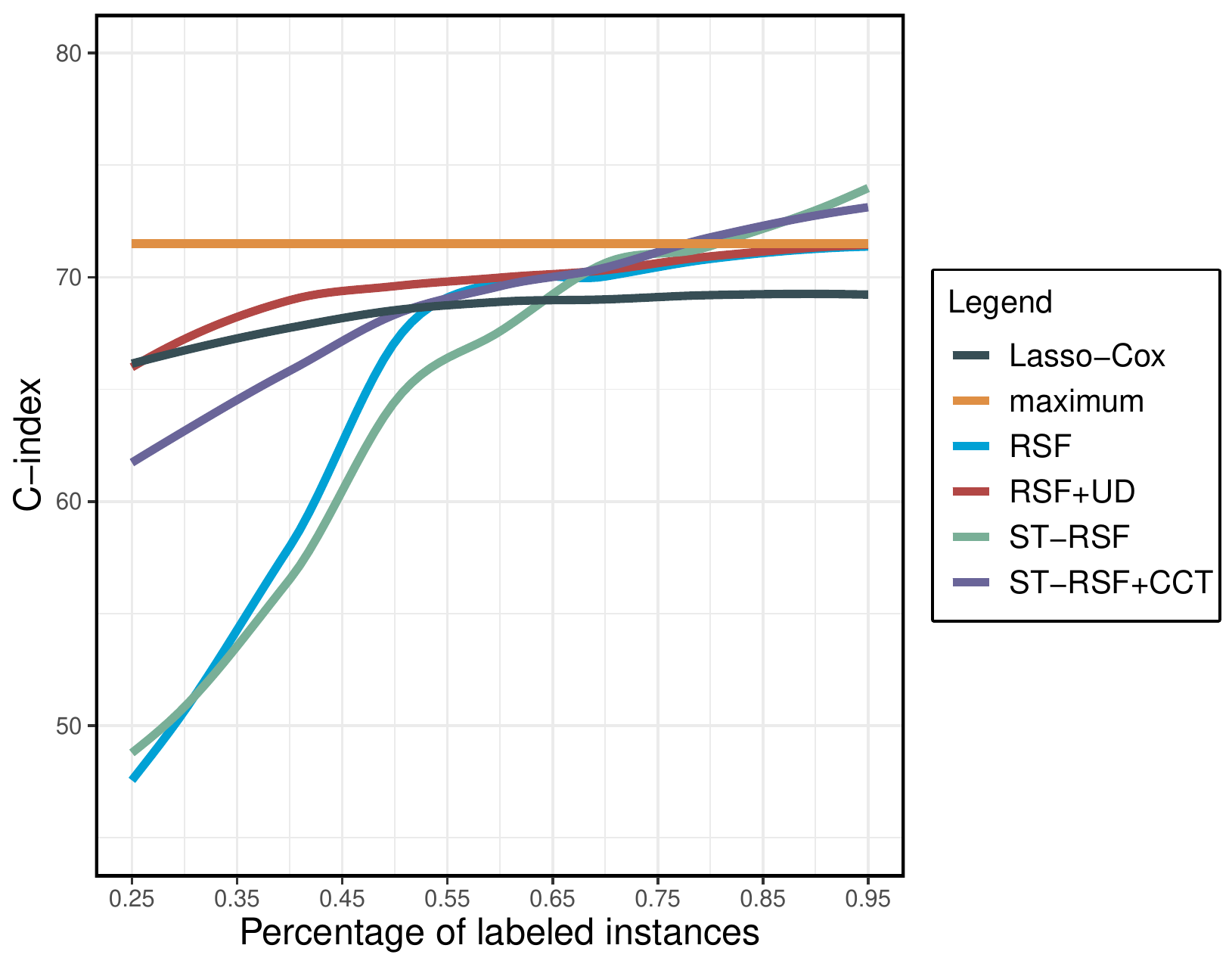}
    \caption{Veteran}
    \label{fig:veteran}
    \end{subfigure}%
        \begin{subfigure}{0.32\textwidth}
          \centering
          \includegraphics[width=\linewidth]{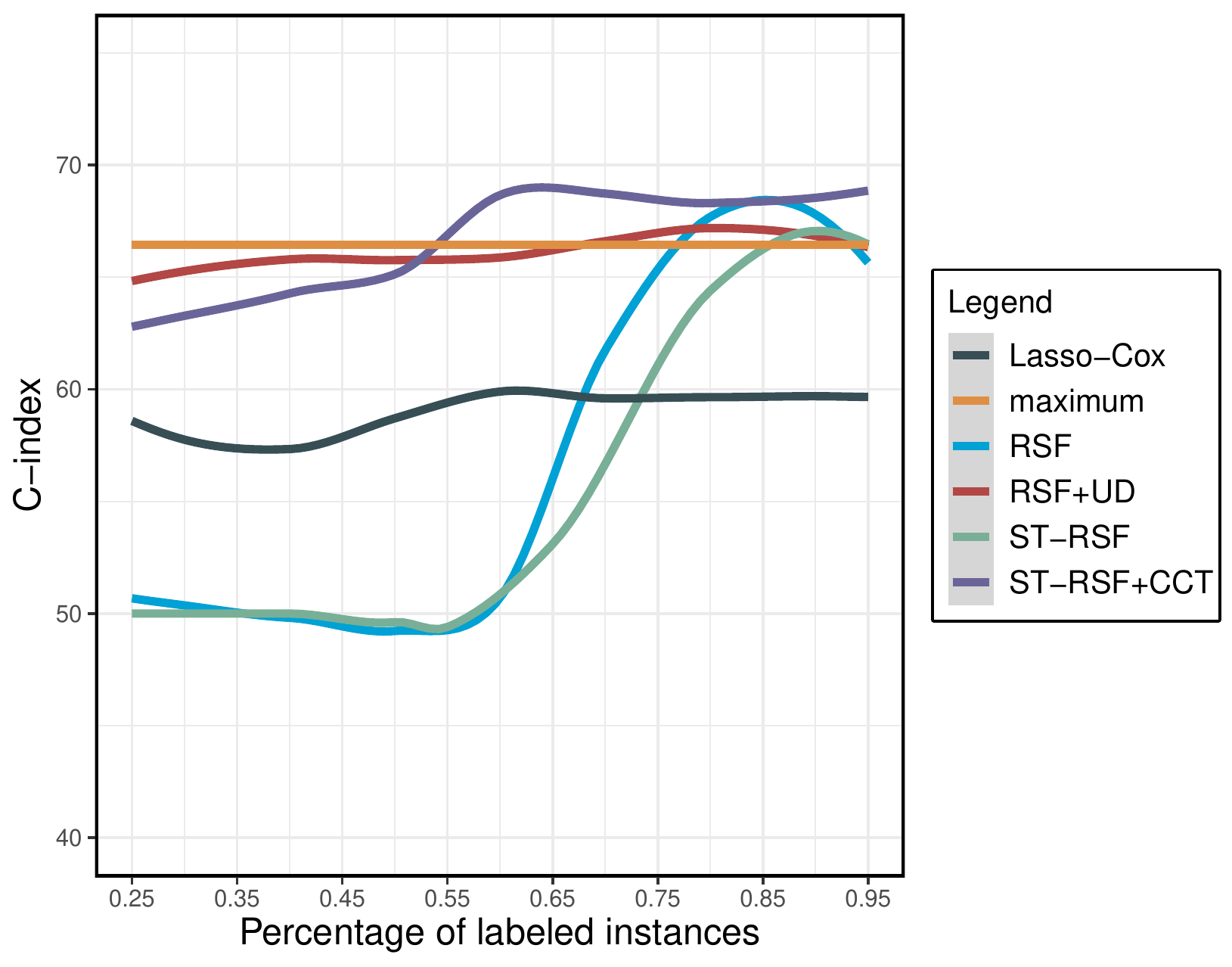}
    \caption{LungBeer}
    \label{fig:LungBeer}
    \end{subfigure}

     \begin{subfigure}{0.32\textwidth}
          \centering
          \includegraphics[width=\linewidth]{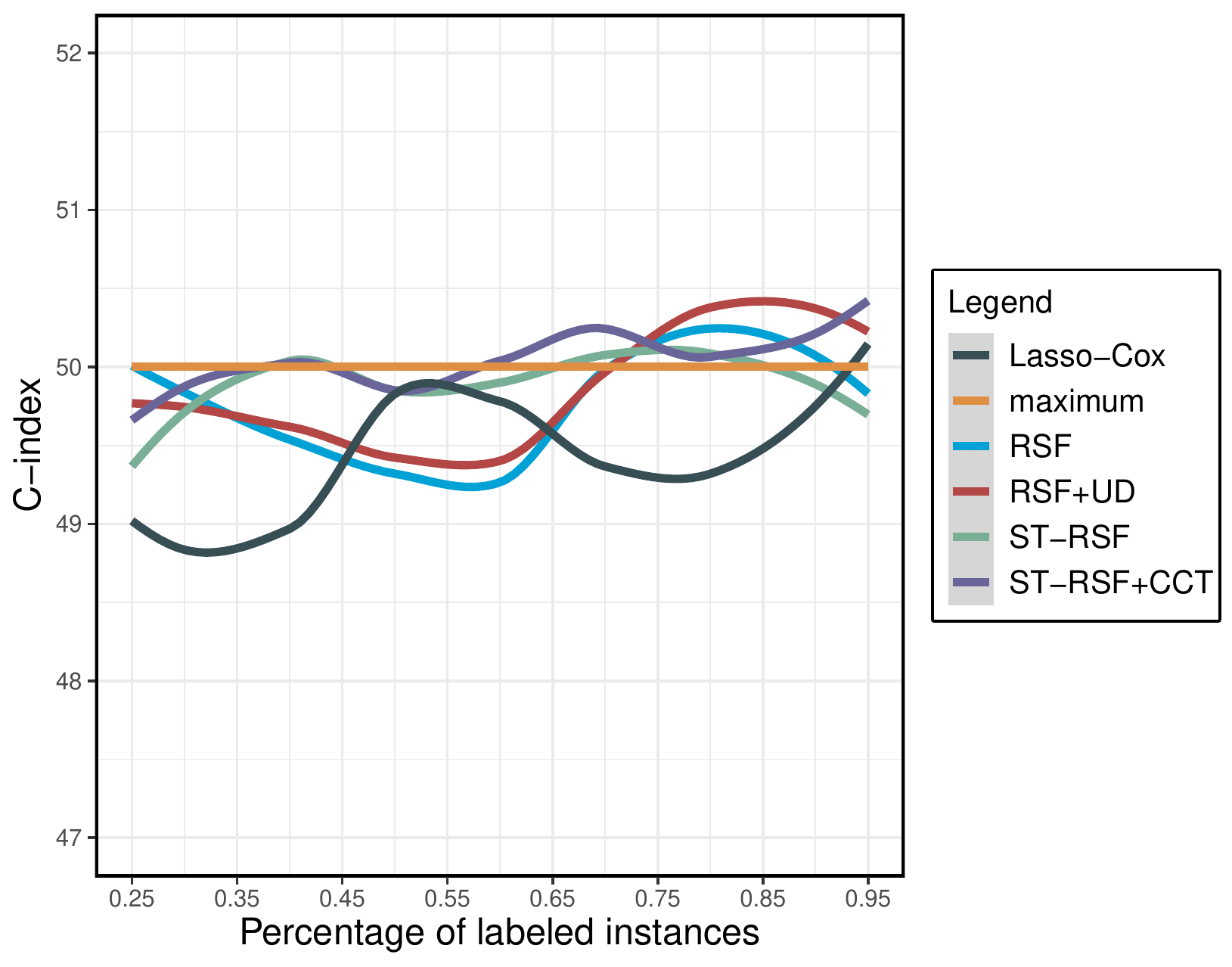}
    \caption{GSE32062}
    \label{fig:GSE32062}
    \end{subfigure}%
     \begin{subfigure}{0.32\textwidth}
          \centering
          \includegraphics[width=\linewidth]{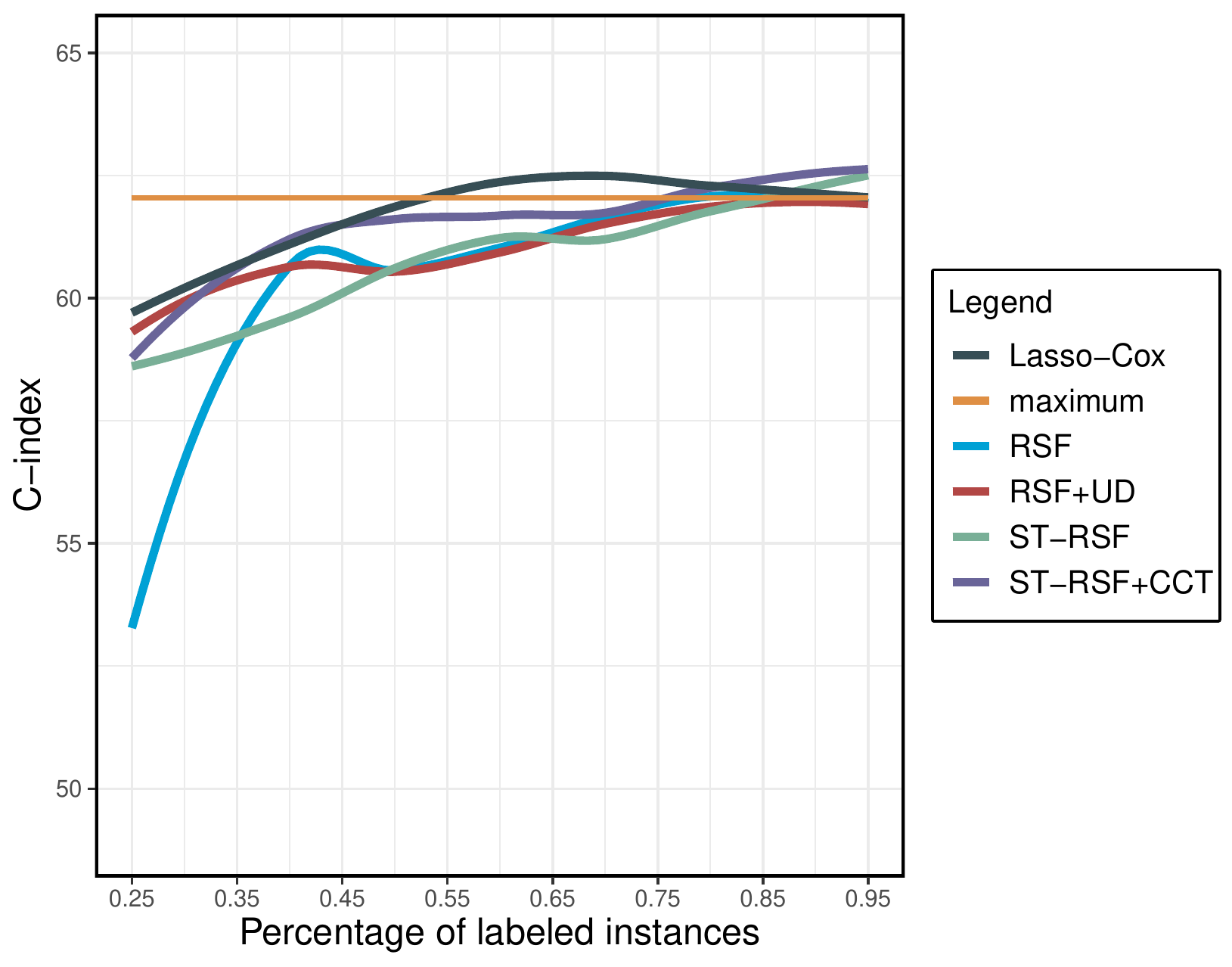}
    \caption{Lung}
    \label{fig:lung}
    \end{subfigure}%
          \begin{subfigure}{0.32\textwidth}
          \centering
          \includegraphics[width=\linewidth]{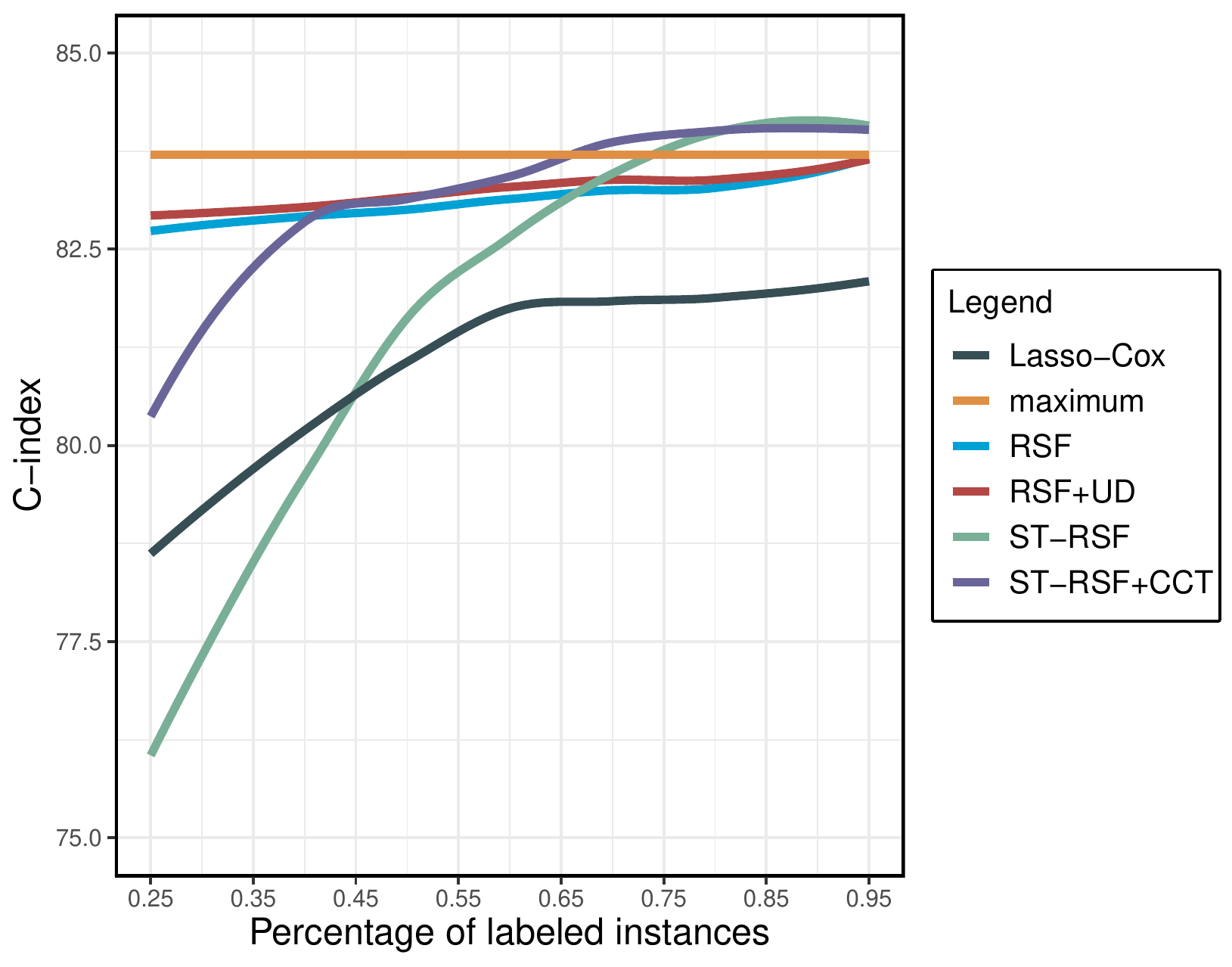}
    \caption{PBC}
    \label{fig:pbc}
    \end{subfigure}

            \caption{Evaluation of the performance of the methods, for different percentages of labeled instances for six datasets.  }
        \label{fig:group2}
\end{figure}

Figures \labelcref{fig:group1,fig:group2} show the performance of the methods, for different percentages of labeled instances for twelve datasets from the fifteen. For each figure, we show six different curves. The blue and dark green curves represent the performance of RSF and Lasso-Cox using only labeled data (Ldata), respectively. The orange line (maximum) shows the performance of RSF using the complete training set as labeled data and is included as a reference to see how much performance we could gain by having access to all (observed or censored) information.
The other three curves represent the proposed approaches.

The figures show that the performance of RSF can indeed be improved by adding unlabeled data to the training set. There are often big performance gains, especially with a lower percentage of labeled instances; however, this improvement does not hold for all datasets and all approaches.

From the figures, we can see that ST-RSF+CCT is the best approach overall, although it often starts in the second or even third position with very few labeled examples. This could be due to a lack of sufficient censored data to guide the augmentation process. Although in three datasets, it starts at a performance lower than RSF, on datasets with a very small number of samples (e.g., Veer, LungBeer, and GSE14764 all with less than 100 observations), ST-RSF+CCT is immediately much better than RSF. Note that 25\% of labeled instances can be as low as 15 labeled examples for the Veer dataset, where ST-RSF+CCT outperforms RSF in the first part of the graph, reaching a C-index level of around 69\%. 

In addition, in several datasets with a high percentage of censored instances (e.g., DBCD, GSE14764, EMTAB, NHANES I, BRCA, and Veer, all with higher than 43\% censoring rate), ST-RSF+CCT is performing as the best method in almost all percentages of labeled instances. 

When comparing the curves for ST-RSF and ST-RSF+CCT, we see in the majority of datasets that either ST-RSF+CCT is on the winning hand over the entire curve, or ST-RSF is better in only some parts. In addition, in most cases, C-index values for ST-RSF fluctuate when changing the number of labeled instances; however, ST-RSF+CCT shows more steady behavior by feeding more labeled instances. Moreover, when comparing the range of C-indices (difference between min and max), ST-RSF varies more dramatically in most experiments; but overall, ST-RSF+CCT acts robustly. This could be due to the fact that for censored instances, ST-RSF+CCT compares the predicted survival time with the censoring time, which results in achieving more confident predictions.

While one would expect the largest gain from using unlabeled data in settings where very few labeled data are available, we see that also considerable improvements can be obtained at the other extreme, where most training instances are labeled and only a small portion, say 5 or 10\%, of unlabeled instances are added. Especially the self-training approaches seem to achieve good results compared to RSF there, although the variability is high. This raises the question of how these techniques would compare to RSF in regular survival analysis tasks (i.e., without an unlabeled set) and can be an interesting direction for future work.

A related observation is that the proposed approaches (especially the semi-supervised ones) are able to beat the `maximum' performance on several occasions. This demonstrates that they are able to select the most reliable instances and leave instances that can harm predictive performance (e.g., noisy instances) out of the training set.

When looking at the RSF+UD curve, we see that it often closely follows the RSF curve for a substantial part of the graph (e.g. for the datasets PBC, GSE32062, DBCD, NHANES I, and BRCA). This is due to the fact that the resulting ensembles are very similar. In fact, the trees generated by RSF are contained in the trees generated by RSF+UD, since the addition of censored data points with event time set to zero does not influence the log-rank splitting criterion, but only alters the size of the trees.

\begin{figure}[b]
    \centering
     \centering\scalebox{.8}{\includegraphics{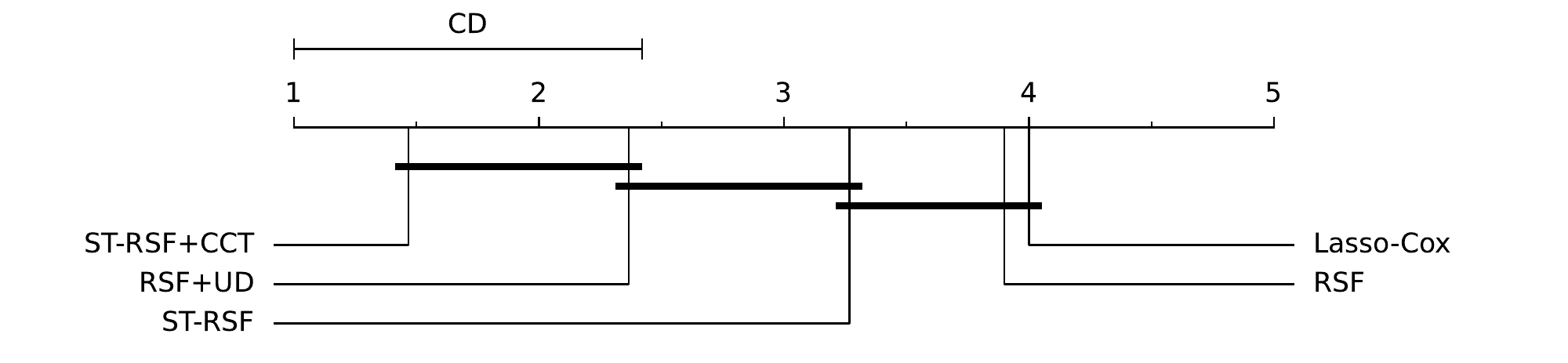}}
    \caption{Results of the Friedman-Nemenyi test of methods ranking. The five methods are compared in terms of their ranking using the evaluation measure, AUC.}
    \label{fig:ranking}
\end{figure}

Since the visual inspection of the figures makes it difficult to draw strong conclusions, we also conducted a more aggregated comparison by comparing the areas under the plotted curves.  Table \ref{table:performance} shows the means and standard deviations of the AUC rate on the datasets, as well as the average accuracy of each algorithm.
As can be seen in Table \ref{table:performance}, all our proposed methods provide better results than RSF for all datasets. More specifically, the third approach (ST-RSF+CCT) outperforms RSF, ST-RSF, and Lasso-Cox and manages to be statistically significantly better according to the Friedman-Nemenyi test (Figure \ref{fig:ranking}) \cite{demvsar2006statistical}\footnote{In a critical distance diagram, those algorithms that are not joined by a line (i.e., their rankings differ more than a critical distance (CD)) can be regarded as statistically significantly different~\cite{demvsar2006statistical}.}. The second best method, on average, is the RSF+UD  variant, which also statistically significantly outperforms RSF and Lasso-Cox, and has a slight, non-significant, margin over ST-RSF.
Furthermore, based on the reported results in Table \ref{table:performance}, in all high-dimensional datasets, either ST-RSF+CCT or ST-RSF are the winning algorithms, meaning that both proposed algorithms are performing better in high-dimensional settings.

\begin{figure}[!t]
    \centering
    \begin{subfigure}{0.5\textwidth}
     \centering
     \includegraphics[width=\linewidth]{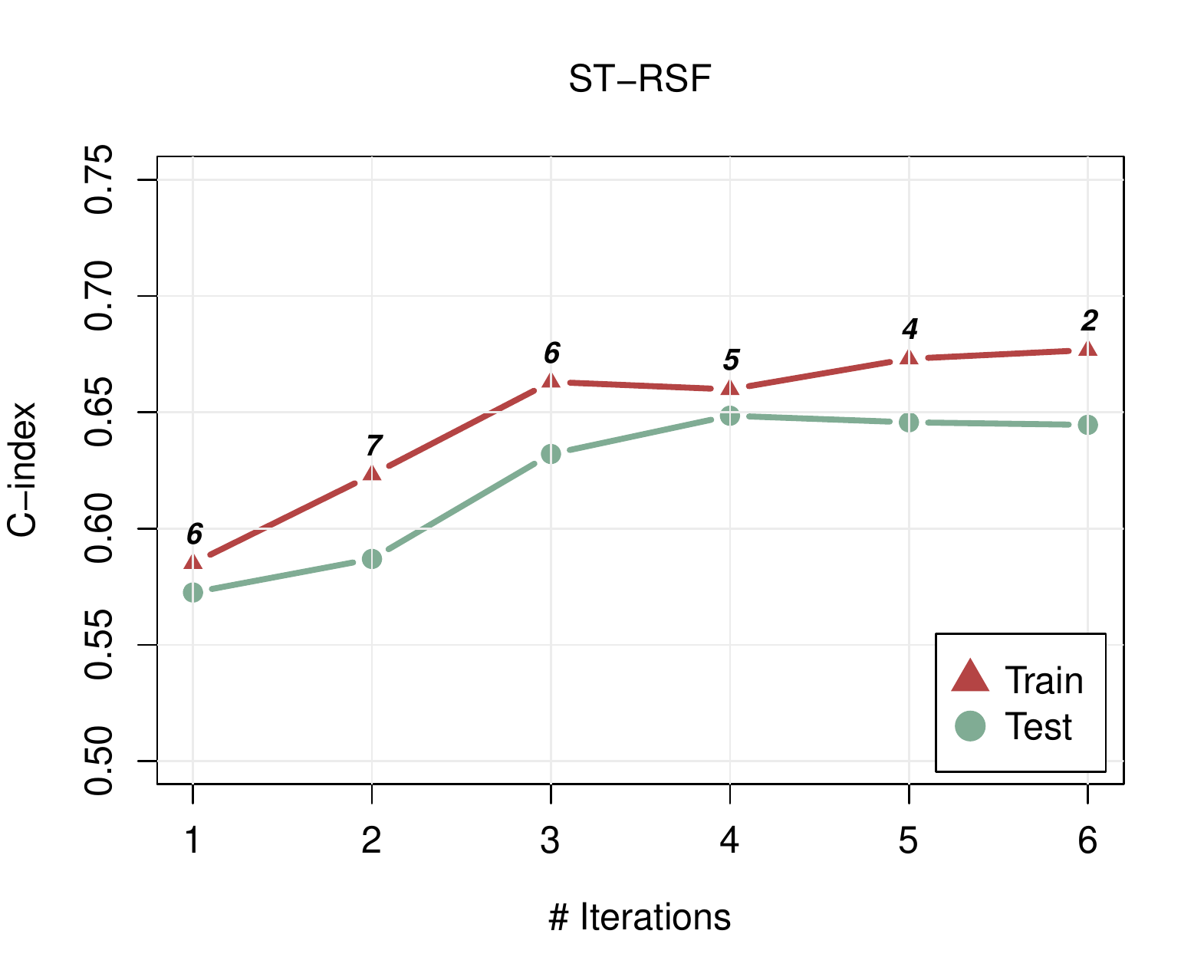}
    \caption{NSBCD}
    \end{subfigure}%
    \begin{subfigure}{0.5\textwidth}
          \centering
          \includegraphics[width=\linewidth]{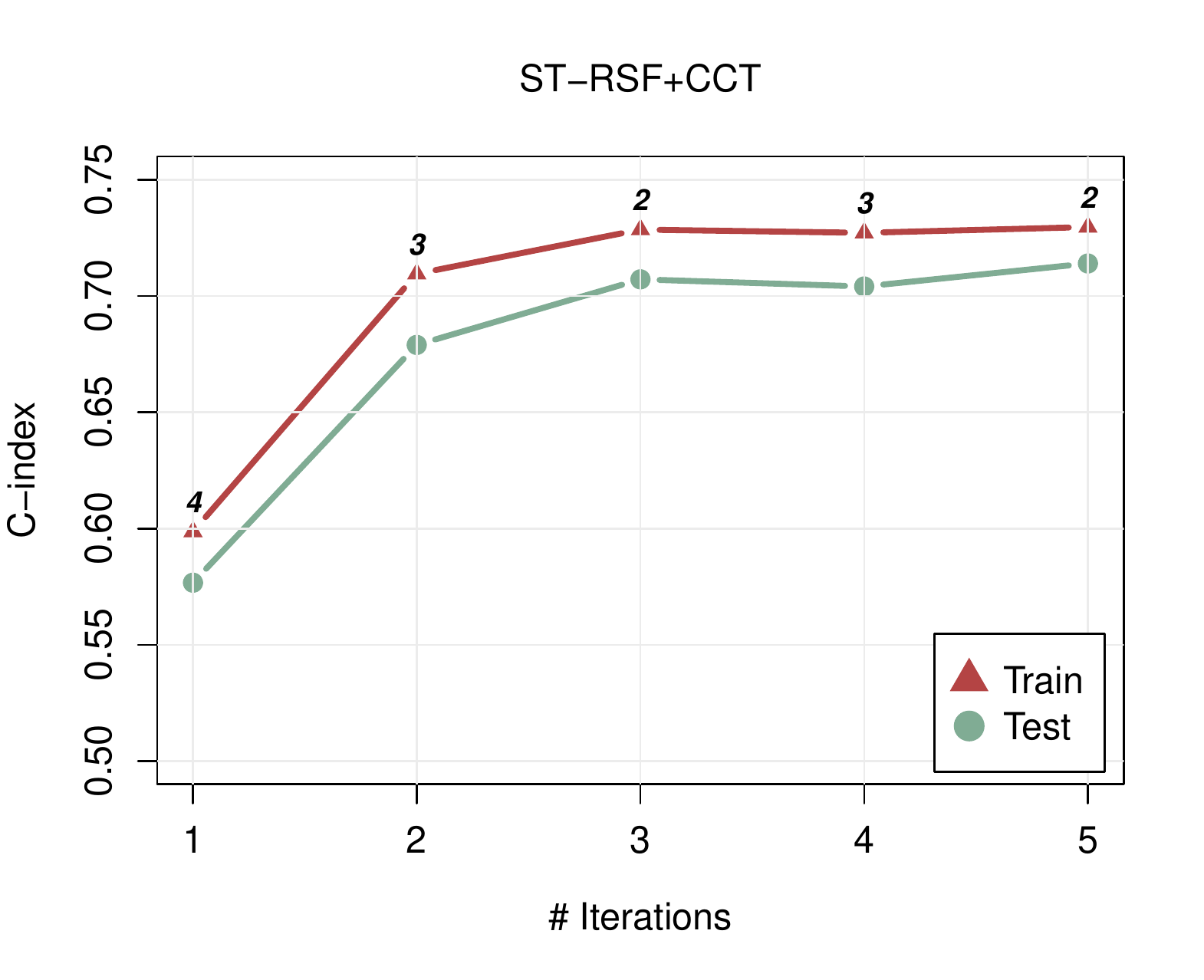}
    \caption{NSBCD}
    \end{subfigure}
    \centering
    \begin{subfigure}{0.5\textwidth}
     \centering
     \includegraphics[width=\linewidth]{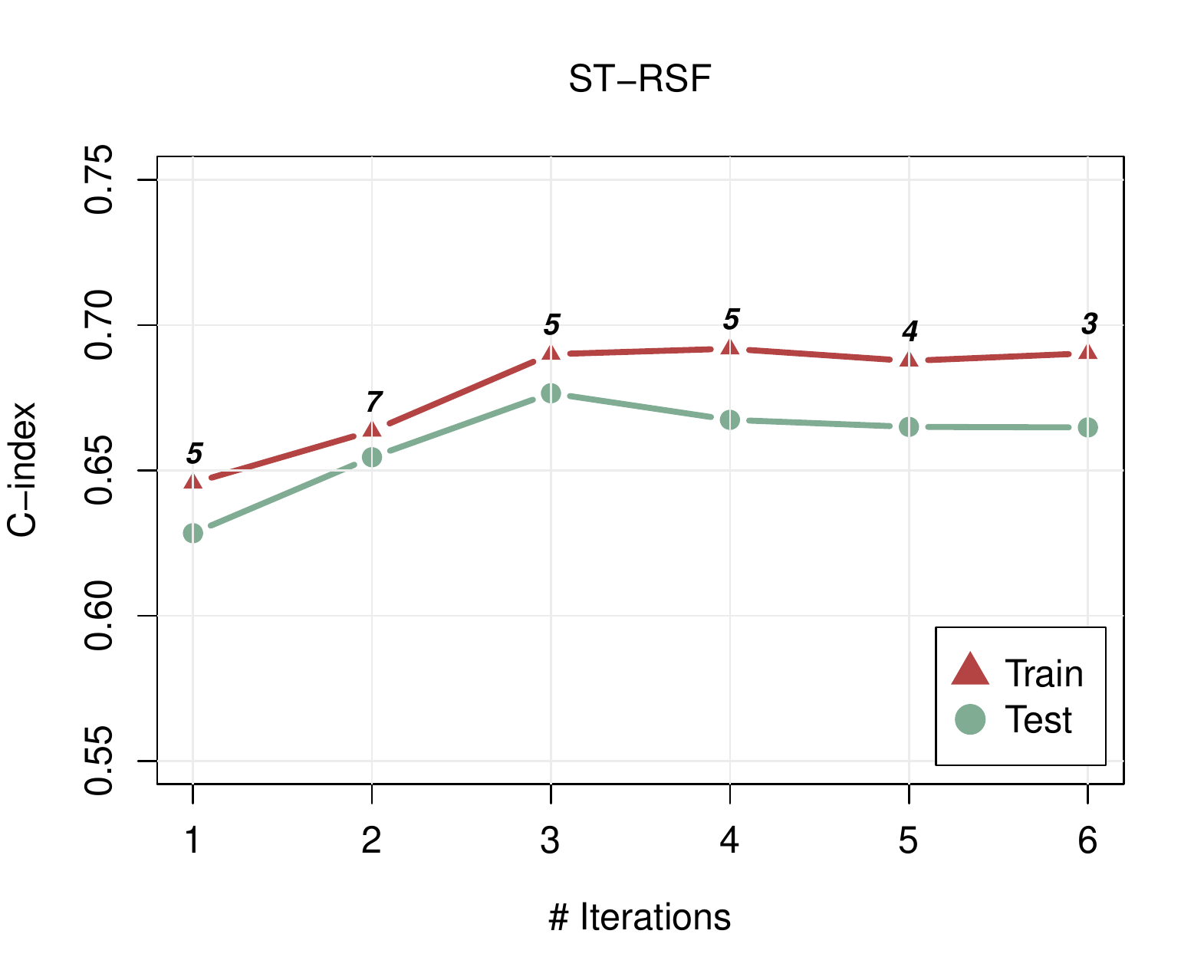}
    \caption{Veteran}
    \end{subfigure}%
    \begin{subfigure}{0.5\textwidth}
          \centering
          \includegraphics[width=\linewidth]{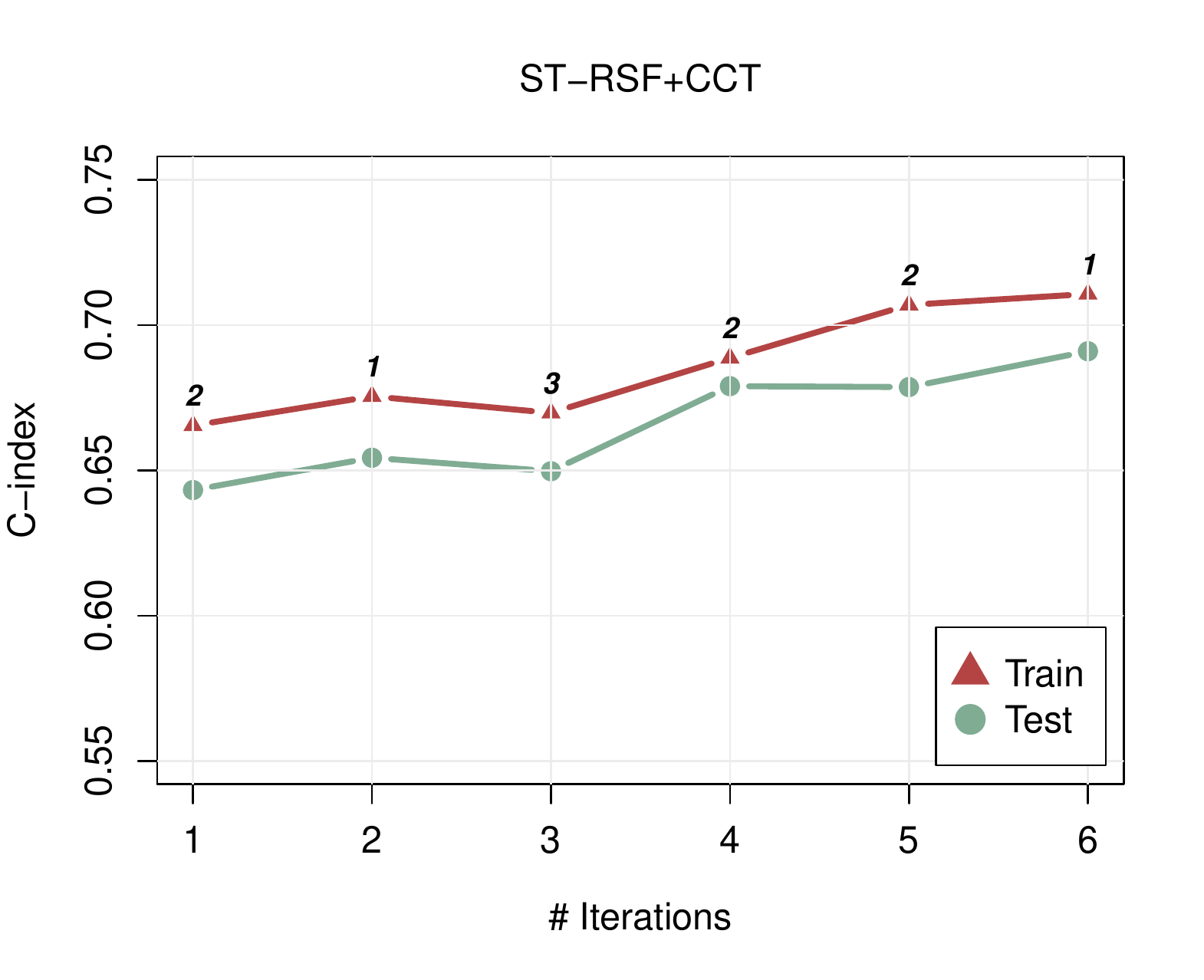}
    \caption{Veteran}
    \end{subfigure}
    
    \caption{Learning curves of the ST-RSF+CCT and ST-RSF methods for NSBCD (figures a and b) and Veteran (figures c and d) datasets. The plots have been shown for 55\% labeled data for both datasets.}
        \label{fig:learning_curve}
\end{figure}

 The use of self-training approaches may raise concerns related to overfitting. Since our algorithms have no tunable hyperparameters, they are not prone to the kind of overfitting that results from the hyperparameter tuning process in other algorithms. 
Moreover, random forests overall are known to be robust to overfitting \cite{breiman2001random}, due to the fact that by increasing the number of trees, the variance of the error gets reduced. 
Nevertheless, we have investigated the learning curve of the ST-RSF and ST-RSF+CCT algorithms on two datasets with 55\% labeled instances (see Figure \ref{fig:learning_curve}). We have chosen NSBCD and Veteran, as they have different censoring rates (67\% versus 6\% censoring rate). In Figure \ref{fig:learning_curve}, the numbers indicated on the training curves show the number of augmented instances in each step. Due to the difference in the augmentation process, in each iteration, ST-RSF+CCT augments fewer instances compared to ST-RSF. For instance, for Veteran, from 42 unlabeled instances, after six iterations (when the stopping conditions hold), ST-RSF has augmented 29 instances, in comparison to 11 for ST-RSF+CCT. The difference in the number of augmentations for ST-RSF and ST-RSF+CCT confirms that the third approach is more conservative and hence leads to less overfitting.

Our findings demonstrate, first, that adding unlabeled data to the training set enhances the performance of the algorithm (cfr.~our first research question), and second, that from the approaches that we have proposed, the self-training technique that uses the information in the censored data points to guide the data augmentation process performs best (cfr.~ our second research question).
The concept of the idea that we proposed could be applied using other base learners and semi-supervised learning strategies, but it remains to be investigated whether the results carry over to other learners.

	\begin{table}[!t]
    \caption{Performance in terms of Area Under the Curve (AUC).
} 
    \centering 
    \begin{adjustbox}{width=0.8\columnwidth,center}
        \def\arraystretch{1}
        \begin{tabular}{c c c c c c} 
            \hline\hline 
            Datasets &  Lasso-Cox & RSF & RSF+UD  & ST-RSF & ST-RSF+CCT \\ [0.5ex] 
            \hline 
Veteran  &$68.4\pm5.1$ & $64.71\pm5.1$      & $\textbf{69.7}\pm5.7$  & $64.33\pm6.07$   & $68.71\pm5.9$\\      
Lung     & $\textbf{61.7}\pm4.64$& $60.55\pm5.3$ & $61.05\pm5.3$  & $60.79\pm5.5$  & $61.52\pm5.02$ \\
PBC      &$81.05\pm3.8$ & $83.12\pm3.4$      & $\textbf{83.24}\pm3.4$  &  $81.67\pm3.6$ & $83.19\pm3.4$\\
BRCA     & $54.81\pm5.6$ & $61.68\pm5.4$      & $61.93\pm5.3$  & $61.1\pm6.5$ & $\textbf{63.14}\pm5.8$\\
 NHANES I & $82.23\pm0.62$ & $82.37\pm0.62$      & $82.37\pm0.62$  &  $82.32\pm0.64$ & $\textbf{82.41}\pm0.7$\\
DrAsGiven &$52.27\pm7.9$ &$ 53.46\pm7.8$      & $55.3\pm8.7$  & $\textbf{56.18}\pm5.2$   & $53.13\pm1.8$ \\
EMTAB386  &$51.54\pm7.6$ & $50.14\pm7.2$      & $50.05\pm8.02$  & $50.12\pm4.02$  & $\textbf{52.83}\pm7.2$ \\
GSE14764  &$51.14\pm14.3$ & $52.14\pm13.7$      & $56.38 \pm16.2$ & $56.65\pm16.1$ & $\textbf{57.52}\pm16.1$ \\
GSE32062  & $49.37\pm5.4$ & $49.76\pm6.3$      & $49.87\pm6.4$  & $49.93\pm6.1$ & $\textbf{50.04}\pm6.2$ \\
NSBCD     & $63.88\pm8.9$ & $62.41\pm7.4$      & $69.34\pm8.1$ & $64.65\pm7.2$  & $\textbf{70.85}\pm7.6$ \\
Veer      & $60.41\pm10.5$ & $54.45\pm15.6$      & $66.49\pm10.1$  & $62.98\pm10.6$  & $\textbf{70.09}\pm9.8$\\
DBCD     & $67.94\pm5.6$ & $71.37\pm5.3$     & $71.38\pm5.3$  &  $70.77\pm5.5$ & $\textbf{73.10}\pm5.7$\\
DLBCL    &$58.12\pm5.2$  & $59.82 \pm5.3$     & $58.94\pm5.3$  &  $\textbf{60.26}\pm5.4$ & $59.82\pm5.3$ \\
LungBeer & $58.94\pm13.2$ & $56.93\pm7.8$      & $66.16\pm12.7$  & $57.21\pm4.8$ & $\textbf{66.67}\pm10.5$ \\
AML      &$59.06\pm8.6$ & $53.3\pm8.7$      & $59.47\pm9.9$  & $ 55.80\pm7.8$ & $\textbf{60.32}\pm9.4$ \\
           
            \hline 
\textbf{Average}      &$61.39$ & $61.01$      & $64.17$  & $62.32$  & $\textbf{64.89}$\\
 [1ex] 
 \hline
        \end{tabular}
    \end{adjustbox}
    \label{table:performance} 
\end{table}

\section{Conclusion} \label{sec:conlcusion }
In this article, we have investigated the inclusion of unlabeled data points in a survival analysis task. More precisely, we have considered learning from data with three degrees of supervision: fully observed, partially observed (censored), and unobserved (unlabeled) data points. To our knowledge, this is a setting that has not been considered before. We have proposed three different approaches for this task. The first approach treats the unlabeled points as censored and applies a standard survival analysis technique. The second one applies a standard semi-supervised wrapper approach on top of a survival analysis task. The third one is an adaptation of the second, which treats the censored instances as unlabeled but manages to exploit the censored information to guide the semi-supervised approach. We have evaluated and compared the proposed approaches on fifteen real-world survival analysis datasets, including clinical and high-dimensional ones. Our results have shown that, first, adding unlabeled instances to the training set improves the predictive performance on an independent test set. Second, the third proposed approach generally outperforms the others due to its ability to integrate partial supervision information inside a semi-supervised learning approach. 

Our findings can be quite helpful, especially in the healthcare area, where studies often require long-term follow-up of patients, which is costly and challenging. For instance, for the prediction of long-term outcomes after hospitalization, our results suggest that the study data could be complemented by additional routinely collected baseline data available in the hospital database management system, from patients matching the inclusion and exclusion criteria, but not included in the follow-up study. 
Moreover, based on the results, our proposed algorithms (ST-RSF+CCT and ST-RSF) perform better in high-dimensional settings (gene-expression datasets) which is a common dataset type in the healthcare area. 

 A limitation of our study is that our experiments assume that the unlabeled set is a random subset of the labeled dataset where the labels have been removed, leading to no trend or bias in the unlabeled set.
When employed in a clinical setting, the unlabeled set should be carefully provided to not incorporate a biased set so that the procedure does not introduce noise through these additive iterations in the algorithm.

\section*{Declarations}
\paragraph{\textbf{Funding.} No funding was received to assist with the preparation of this manuscript}
\paragraph{\textbf{Conflicts of interest/Competing interests.} The authors have no conflicts of interest to declare that are relevant to the content of this article}
\paragraph{\textbf{Ethics approval.} Not applicable}
\paragraph{\textbf{Consent to participate.} No tests, measurements, or experiments were performed on humans as part of this work}
\paragraph{\textbf{Consent for publication.} The authors have agreed to submit it in its current form for consideration for publication in Journal}
\paragraph{\textbf{Availability of data and material.} All of the datasets used in this article are publicly available and have been referenced}
\paragraph{\textbf{Code availability.} The source code will be publicly available}
\paragraph{\textbf{Authors' contributions.} 
\textbf{Fateme Nateghi Haredasht:} Investigation, Methodology, Software, Writing, Original draft. 
\textbf{Celine Vens:} Supervision, Conceptualization, Methodology, Writing, Review and editing} 

\section*{Acknowledgements}
This work was supported by KU Leuven Internal Funds (grant 3M180314). The authors also acknowledge the Flemish Government (AI Research Program).
\clearpage
\bibliography{mybibfile}

\end{document}